\documentclass[final]{elsarticle}

\journal{Information Sciences. Accepted for publication.}

\usepackage{graphicx}
\usepackage{algorithm}
\usepackage{algpseudocode}
\usepackage{epstopdf}
\usepackage[utf8]{inputenc}

\usepackage{todonotes}

\usepackage{amssymb}
\usepackage{amsmath,bm}
\usepackage{url}
\usepackage{subcaption}
\usepackage{lineno}
\usepackage{xcolor}
\usepackage{ulem}
\usepackage{setspace}
\usepackage{eqnarray}
\usepackage{balance}
\usepackage[acronym]{glossaries}
\usepackage{algorithm}
\usepackage{multirow}
\usepackage{soul}

\DeclareMathOperator*{\argmin}{argmin}

\begin{document}

\begin{frontmatter}


\title{Incremental and Decremental Fuzzy Bounded Twin Support Vector Machine}


 \author[label1,label2,label3]{Alexandre R. Mello\corref{}}\ead{alexandre.reeberg@posgrad.ufsc.br} \ead[url]{https://orcid.org/0000-0003-3130-5328}
\author[label2]{Marcelo R. Stemmer}
\ead{marcelo.stemmer@ufsc.br}
\author[label1]{Alessandro L. Koerich}
\ead{alessandro.koerich@etsmtl.ca}
\address[label1]{École de Technologie Supérieure - Université du Québec, 1100 Notre-Dame West, Montréal, QC, H3C 1K3, Canada.}
\address[label2]{University of Santa Catarina, Campus Reitor João David Ferreira Lima, Trindade, Florianópolis, SC, 88040-900, Brazil.}
\address[label3]{SENAI Innovation Institute of Embedded Systems, Avenida  Luiz Boiteux Piazza, 574 - Cond. Sapiens Parque - Canasvieiras - Florianópolis, SC, 88054-700, Brazil.}

\begin{abstract}
In this paper, we present an incremental variant of the Twin Support Vector Machine (TWSVM) called Fuzzy Bounded Twin Support Vector Machine (FBTWSVM) to deal with large datasets and to learn from data streams. We combine the TWSVM with a fuzzy membership function, so that each input has a different contribution to each hyperplane in a binary classifier. To solve the pair of quadratic programming problems (QPPs), we use a dual coordinate descent algorithm with a shrinking strategy, and to obtain a robust classification with a fast training we propose the use of a Fourier Gaussian approximation function with our linear FBTWSVM. Inspired by the shrinking technique, the incremental algorithm re-utilizes part of the training method with some heuristics, while the decremental procedure is based on a {\color{black}scoring} window. The FBTWSVM is also extended for multi-class problems by combining binary classifiers using a Directed Acyclic Graph (DAG) approach. Moreover, we analyzed the theoretical foundation's properties of the proposed approach and its extension, and the experimental results on benchmark datasets indicate that the FBTWSVM has a fast training and retraining process while maintaining a robust classification performance.
\end{abstract}
\end{frontmatter}

\section{Introduction}\label{sec:introduction}

Classical machine learning approaches, in which all data is simultaneously accessed, do not meet the requirements to deal with the scenario in which training data is partially available at a time or where the amount of data is so large that it does not fit into the memory or the storage of a single machine. Incremental or online learning is an approach to tackle problems in which only a subset of the data is considered at each step of the learning process, or when the dataset is too large to be processed at once \cite{Khemchandani2009}. From the computational point of view, incremental learning has three goals:
(i) transform previously learned knowledge to \textcolor{black}{currently} received data to facilitate learning from new data; (ii) accumulate experience over time to support the decision-making process; (iii) and achieve global generalization through learning to accomplish goals. Incremental learning often also refers to on-line learning strategies with limited memory resources, relying on creating a compact memory model that represents the already observed data but providing accurate results for all relevant settings.

Losing et al.~\cite{Losing2018} evaluated the most common algorithms of incremental learning on diverse datasets, and the conclusion is that the Support Vector Machines (SVMs) are usually the highest accurate models. However, such accuracy is at the expense of the most complex model besides many other shortcomings. SVMs \textcolor{black}{are appropriate} to tackle two-class classification problems by solving a complex Quadratic Programming Problem (QPP) that determines a unique global hyperplane in the input space that maximizes the separation between the classes \cite{Cortes1995a}. However, it requires large memory and a high CPU power since the computational complexity of the SVM for $n$ data points is $O(n^3)$, which makes it impractical for large datasets. To circumvent this problem, one may use the incremental version of SVM or its variants, that learns from new data by discarding past data points excepting the support vectors (SVs), i.e., the new data is used to retrain the model together with the current SVs \cite{Khemchandani2009,Cauwenberghs2000}.

The Incremental SVM (ISVM) proposed by Cauwenberghs and Poggio \cite{Cauwenberghs2000} is an exact solution to the problem of on-line SVM that updates the optimal solution of the SVM by adding or removing one training data point. The bottleneck of the ISVM is that the computational complexity of a minor iteration of the algorithm is quadratic in the number of training data points learned so far. Therefore, the actual runtime depends on the balance between memory access and arithmetic operations in a minor iteration 
The LASVM \cite{Bordes2005a} is an on-line kernel classifier that relies on the soft-margin SVM formulation to handle noisy data. The iterations are similar to the sequential minimization optimization (SMO) algorithm but with a different search strategy. Furthermore, it introduces \textcolor{black}{an} SV removal step, where it removes the vectors collected in the current kernel expansion during the on-line process. The iterations run in epochs, where each epoch sequentially visits all the randomly shuffled training data points, and the stopping criteria is a pre-defined number of epochs. \textcolor{black}{We can use a multiple number of epochs} as a stochastic optimization algorithm in the off-line training, and a single epoch in the on-line step. The computational cost of the LASVM is $O(p\times nSV\times i)$, where (\textit{nSV}) is the number of SVs, \textit{i} is the number of on-line iterations, and $p$ scales no more than linearly to the amount of training data points, which makes the training process faster than the ISVM. Empirical results suggest that using a single epoch yields to misclassification rates comparable with the SVM. Despite the effectiveness of the ISVM and the LASVM, both methods still need to deal with one large QPP, requiring large memory storage and CPU processing time on training and update steps. 
Mangasarian et al.~\cite{Mangasarian2006} introduced the Generalized Eigenvalue Proximal SVM (GEPSVM) that generates two non-parallel hyperplanes for a two-class problem. Thus, it solves two smaller QPPs instead of a single complex QPP, laying each class data point in the proximity of a hyperplane, which reduces the complexity compared to the SVM. Jayadeva et al.~\cite{Jayadeva2007} proposed the Twin Support Vector Machine (TWSVM), which also solves a pair of QPPs where the data points of one class provide constraints to the other QPP and vice versa \cite{Tomar2015,Ding2014}. The TWSVM classifies the data points of two classes using two non-parallel hyperplanes with a complexity of $O(2$$\times$$(n/2)^3)$, which is four times lower than \textcolor{black}{an} SVM. Twin-based models are mathematically smaller than the SVM, and they require low memory storage and CPU processing time.

Based on the TWSVM, several variants and solvers have been proposed \cite{Tomar2015,Jayadeva2017a}. 
Yuan-HaiShao et al.~\cite{Yuan-HaiShao2011} suggested the Twin Bounded SVM (TBSVM) that includes adherence to the structural risk minimization principle, so the dual formulation (whose inverse is guaranteed) can be solved by successive over-relaxation (SOR) methodology. The Improved TWSVM (ITWSVM) \cite{Tian2014} uses a different representation from the TBSVM that leads to a different Lagrangian function for the primal problem and different dual formulations. The ITWSVM does not need to compute the inverse of large matrices before training and can be solved by the SOR or the SMO. However, the matrices in the dual form must involve all the data points from both classes, which makes the dual QPPs larger than the TWSVM. Khemchandani et al.~\cite{Khemchandani2008} proposed a novel fuzzy TWSVM that assigns a fuzzy weight to each data point to mitigate the effect of outliers and improve accuracy. Gao et al.~\cite{Gao2015} proposed a coordinate descent fuzzy TWSVM, assigning a membership function to mitigate the effect of noisy data points, and solving the QPPs with a coordinate descent with shrinking by active set. Other variants or extensions are the Least Square TWSVM (LS-TWSVM) \cite{ArunKumar2009} that solves the primal problems of the TWSVM, and the $\nu$-TWSVM \cite{Peng2010} where the $\nu$ parameter controls the bounds of the fractions of the SVs and the error margin.

Considering the TWSVM and its variations, Khemchandani et al.~\cite{Khemchandani2009} introduced the incremental TWSVM (I-TWSVM), which uses the concept of margin vectors and error vectors to select new data points to update the classifier. It learns from new data by retraining the model while discarding past data points except for the previous SVs and erroneous classified data points from the training dataset. However, for each new data point, both models need to be \textcolor{black}{rebuilt entirely}. Hao et al.~\cite{Hao2014} proposed a fast incremental TWSVM that uses a distance-based strategy to determine if a new data point is above a pre-defined threshold. It selects the \textcolor{black}{crucial} data points that are near the proximal hyperplane from the current training set and keep data points that are not near the proximal hyperplane from the new training set. In each iteration, it retrains the model considering the previous SVs and the new data points (there is no decremental step). The on-line Twin Independent SVM (OTWISVM) \cite{Alamdar2016} uses a modified Newton method to build a decision function via a subset of data points seen so far for each class separately (called basis). The basis vectors are found (or added during the on-line procedure) during iterative minimization by checking if a new data point is linearly independent in the feature space from the current basis. The basis size is limited, so it does not grow linearly with the training set. The OTWISVM does not have a decremental step, and as it utilizes a modified Newton solver, it needs to calculate the inverse of the Hessian on every update, making the method unfeasible to deal with high-dimensional datasets. Besides improving the model with new data, it is also \textcolor{black}{essential} to have a decremental procedure to prevent the model from growing indefinitely. Despite the update strategy be closely related to the model formulation, there are many alternatives \textcolor{black}{to} choosing the SVs to be removed, such as the time-window proposed by Fung et al.~\cite{Fung2002}, the concept of informative margin vectors and error vectors \cite{Cauwenberghs2000}, or decay coefficients \cite{Tveit2003}.

Finally, to unleash the full potential of the incremental SVM, it is necessary to adapt it to deal with non-linear problems using the kernel trick. However, conventional kernel approaches struggle to deal with large datasets due to the storage and computational issues in handling large kernel matrices. A feasible solution {\color{black}is using} kernel approximations such as: exploiting low-rank approximation of the kernel matrix\textcolor{black}{,} reducing the kernel space definition\textcolor{black}{,} or exploiting a randomized kernel space definition. 
Random Fourier approximations (RF) provide an efficient and elegant methodology \cite{Rahimi2008}. The Fourier expansion generates features based on a finite set of random basis projections with inner products that are the kernel Monte Carlo approximations \cite{Li2010}. Fourier features are applicable to translation-invariant kernels, so  {\color{black} they can} approximate the Gaussian kernel. Rahimi et al.~\cite{Rahimi2008} use RF to map the input data to a randomized low-dimensional feature space providing convergence bounds to approximate various radial basis kernel. Le et al.~\cite{Le2014a} proposed an RF-based approximation called \textit{Fastfood}, which requires a smaller computation and memory storage than Random Kitchen Sinks \cite{Rahimi2009} to obtain an explicit function space expansion. 

Although many efforts have been made, the incremental SVM approaches still have several shortcomings, such as the impossibility of endless learning, high model complexity, high training time, high complexity of hyper-parameter optimization, adaptability to concept drifts, among others. In this paper, we propose a novel incremental and decremental variant of the TWSVM called Fuzzy Bounded Twin Support Vector Machine (FBTWSVM) that overcomes many of the shortcomings of the current approaches. The FBTWSVM combines fast training and an incremental procedure (with the ability to handle noisy data) without weakening the accuracy when updated. The proposed approach can continuously integrate new information into already-built models\textcolor{black}{,} \textcolor{black}{with the characteristics of being adherent} to the structural risk minimization principle \cite{Yuan-HaiShao2011}, and  \textcolor{black}{using} the dual coordinate descent (DCD) algorithm with active shrinking \cite{Tian2014,Khemchandani2008,Gao2015,Gao2017} to create the off-line classifier. The incremental and decremental strategies are based on the DCD with shrinking, exploiting the relevance of each support vector. Moreover, we propose the use of our linear formulation with a kernel approximation to speed up training and classification while maintaining the non-linearity. Finally, the FBTWSVM is extended to multiclass problems using a strategy based on the Directed Acyclic Graph (DAG). The experimental results on benchmarking datasets have shown that the proposed approach achieves accuracy comparable to the exact solution besides being faster to integrate new information and to discard outdated information into the already-built models.

This paper is organized as follows. Section~\ref{se:Basic} presents the definitions and notations, introduces the Twin SVM, and presents the fuzzy SVM with the respective membership function and the kernel approximation. Section~\ref{se:FBTSVM} presents {\color{black} the }proposed formulation for the FBTWSVM (both linear and non-linear versions), {\color{black}and} the solving method with implementation details. We also extend our formulation to multiclass {\color{black}problems}. In Section~\ref{sec:inc_dec}\textcolor{black}{,} we present the incremental and decremental procedures. In Section \ref{se:Experimental}\textcolor{black}{,} we {\color{black}present} our experimental procedure and the experimental results. \textcolor{black}{We present the} conclusions and perspectives for future work in the last section.

\section{Basic Concepts}
\label{se:Basic}
We use the following definitions and notations throughout the paper. The problems are in a $n-$dimensional space $\mathcal{R}^n$. We denote the training data as $D$$=$${(\bm{x_i},y_i)|i=1,2,\dots,l}$, where $\bm{x_i}\in\mathcal{R}^n$ is an input data point, and $l$ is the number of data points, with the corresponding label $y_i\in\{1,2,\dots,u\}$ where $u$ is the number of classes. We adopt the definition of incremental learning proposed by Losing et al.~\cite{Losing2018} as an algorithm that generates on a given stream of training data $\bm{x}_1,\bm{x}_2,\dots,\bm{x}_t$ a sequence of models $\bm{h}_1,\bm{h}_2,\dots,\bm{h}_t$, where $(\bm{h}_i$$:$$\mathcal{R}^n|i=1,2,\dots,l)$ is a model function solely depending on $\bm{h}_{i-1}$ and the recent $p$ data points $\bm{x}_i,\dots,\bm{x}_{i-p}$ with $p$ being strictly limited. The approach used to deal with multiclass problems is the DAG, where it is necessary to create $2u-$$1$ binary problems. For each binary problem we assign either a positive or a negative label $y_i\in\{+1,-1\}$. Therefore, the training set $D$ is divided into the $l_+ \times n$ dimensional matrix $X_+$ and $l_-\times n$ dimensional matrix $X_-$ for positive and negative labels respectively, where $l_+$ and $l_-$ denote the number of data points from each label. We define the aggregation per binary problem as $X$$=$$[X^\top_+ X^\top_-]$, and it denotes all input data points from both classes.

\subsection{The Twin SVM (TWSVM)} \label{se:TWSVM}
\begin{figure}[hbtp!]
	\centering
	\includegraphics[width=0.55\linewidth]{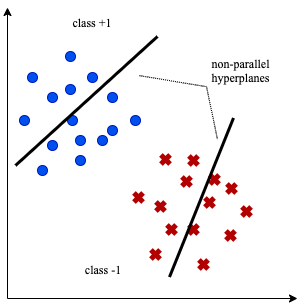}
	\caption{Binary classification using the TWSVM, inspired by \cite{Tomar2015}.}
	\label{fig:TWSVM}
\end{figure}

The TWSVM \cite{Jayadeva2007} generates two non-parallel hyperplanes such that each hyperplane is closer to one class and is as far as possible from the other class, \cite{Tomar2015,Jayadeva2017} as shown in Figure~\ref{fig:TWSVM}. The two non-parallel decision planes are defined as:

\begin{equation}\label{eq:Twinplanes}
	\bm{\omega^\top_+} \bm{x}+b_+=0 \;\; \text{and}\;\; \bm{\omega^\top_-} \bm{x} +b_-=0
\end{equation}

\noindent where $\bm{\omega_+},\bm{\omega_-}$$\in\mathcal{R}^n$ indicates normal vectors to the hyperplane, and $b_+,b_-\in\mathcal{R}^n$ are the bias terms.

\textcolor{black}{The following pair of primal optimization problems is the setup to build the decision planes, as the soft margin hyperplane can handle non-linearly separable data:}

\begin{equation}\label{eq:TWSVM+}
	\begin{aligned}
		&	\min_{ \bm{\omega_+},b_+,\xi_-} \;\; \frac{1}{2} ||X_+ \; \bm{\omega_+} + \bm{e_+} \; b_+||^2 + C_1 \; \bm{e^\top_-} \; \xi_- \\
		&	\;\; \text{s.t. } y_- (X_- \; \bm{\omega_+} + \bm{e_-} \; b_+)+\xi_- \leq \bm{e_-}, \;\; \xi_-\geq 0
	\end{aligned}
\end{equation}    
and
\begin{equation}\label{eq:TWSVM-}
	\begin{aligned}
		&	\min_{\bm{\omega_-},b_-,\xi_+} \;\; \frac{1}{2} ||X_- \; \bm{\omega_-} + \bm{e_-} \; b_-||^2 + C_2 \; \bm{e^\top_+} \; \xi_+ \\
		& \;\;	\text{s.t. } y_+(X_+ \; \bm{\omega_-} + \bm{e_+} \; b_-)+\xi_+ \leq \bm{e_+}, \;\; \xi_+\geq 0
	\end{aligned}
\end{equation}        

\noindent where $C_1>0$  and $C_2>0$ are the penalty factors that trade-off the complexity and data misfit between the minimization of the two terms in the objective function, $\xi_+$ and $\xi_-$ denote the slack variable vectors (the deviation from the margin that allows subsets of misclassification error for positive and negative classes respectively), $\bm{e_-}$, $\bm{e_-}$ correspond to unit row vectors with their dimensions exact to data point size in each class used for mathematical purpose only, $y_+$ and $y_-$ are $+1$ and $-1$ respectively. 

In each QPP (Equations \ref{eq:TWSVM+} and \ref{eq:TWSVM-}) the objective function corresponds to a particular class and the constraints are set by the data points of the opposite class. \textcolor{black}{The first term of both objective functions aims to minimize the sum of squared distances between the hyperplane and the points of one class, which tends to keep the hyperplane close to the points of such a class. On the other hand, the second term of both objective functions aims to minimize the misclassification due to points belonging to the other class. The constraints require the distance between the hyperplane and the points of the other class is at least 1, and a set of error variables is used to measure the error wherever the hyperplane is closer than the minimum distance \cite{Jayadeva2007}. }Assuming that the TWSVM is split into two QPPs of size $n/2$, and that the complexity of the original SVM is less or equal to $n^3$, the TWSVM is approximately four times faster than the original SVM ($2\times(n/2)^3 = n^3/4$)\cite{Jayadeva2017}. After solving Equations~\ref{eq:TWSVM+} and~\ref{eq:TWSVM-} for $(\bm{\omega^*_}+,b^*_+)$ and $(\bm{\omega^*_-},b^*_-)$, respectively, we can classify a new data point $\bm{x}$ by:

\begin{equation}
	\label{eq:Classif}
	f(x)=\argmin_\pm \frac{|\bm{\omega^{*\top}_\pm} \bm{x} +b^*_\pm|}{||\bm{\omega^*_\pm}||}
\end{equation}

\noindent and choose either $+1$ or $-1$ according to the lowest value of Equation~\ref{eq:Classif}, \textcolor{black}{i.e., we classify a new data point $\bm{x}$ depending on which of the two hyperplanes given by Equation~\ref{eq:Twinplanes} it lies closest.} We can write Equations~\ref{eq:TWSVM+} and~\ref{eq:TWSVM-} as an unconstrained problem using Lagrangian multipliers. The dual formulation of the linear TWSVM for Equation~\ref{eq:TWSVM+} is:

\begin{equation}
	\label{eq:dualTWSVM1}
	\begin{aligned}
		&\max_{\bm{\alpha}} \;\;\; \bm{e^\top_-} \bm{\alpha} -\frac{1}{2}\bm{\alpha^\top} H_- (H^\top_+ H_+)^{-1}H^\top_- \bm{\alpha} \\
		&\;\;	\text{s.t.  } 0\leq \bm{\alpha} \leq C_1
	\end{aligned}
\end{equation}

\noindent where $H_+$$=$$[X_+,e_+]$, $H_-$$=$$[X_-,e_-]$, $||$$\cdot$$||$ denotes the L2 norm, and $\bm{\alpha}$$=$$(\alpha_1,\dots,\alpha_m)^\top$ is the vector of Lagrangian multipliers. In a similar manner we can write the dual formulation for Eq~\ref{eq:TWSVM-} as:
\begin{equation}\label{eq:dualTWSVM2}
	\begin{aligned}
		&	\max_{\bm{\nu}} \;\;\; \bm{e^\top_+}\; \bm{\nu} -\frac{1}{2}\bm{\nu^\top} H_+ (H^\top_- H_-)^{-1}H^\top_+ \bm{\nu} \\
		&\;\;	\text{s.t.  } 0\leq \bm{\nu} \leq C_2
	\end{aligned}
\end{equation}

\noindent where $\bm{\nu}$$=$$(\nu_1,\nu_2,\dots,\nu_{m})$ is the vector of Lagrangian multipliers. For more detail on the dual formulation one may refer to \cite{Jayadeva2007,Jayadeva2017}. Once we solve dual problems for $\bm{\alpha}$ and $\bm{\nu}$, we can get the vectors $[\bm{\omega_+},b_+]^\top$ and $[\bm{\omega_-},b_-]^\top$. Thus, the separating hyperplanes are given by:

\begin{equation}
	\bm{x^\top} \bm{\omega_+} + b_+=0 \;\; \text{and}\;\; \bm{x^\top} \bm{\omega_-} + b_- =0
\end{equation}

During testing, a new data point is assigned to the closest hyperplane regarding the two classes by:

\begin{equation}
\color{black}
	class(\bm{x})=\argmin_{i=\{-1,+1\}}d_i(\bm{x})
\end{equation}

\noindent where

\begin{equation}
	d_i(x)=\frac{|\bm{x^\top} \bm{\omega_i} + b_i|}{||\bm{\omega_i}||}
\end{equation}

\subsection{The Fuzzy SVM}
\label{se:Fuzzy}
{\color{black}The presence of outliers in the training dataset may affect both the standard SVM and the TWSVM.} The Fuzzy SVM introduced by Lin et al.~\cite{Chun-FuLin2002} uses the fuzzy theory to reduce the effect of outliers by applying a membership to each data point. Fuzzy numbers, denoted as $s_i$, are assigned to each input data point to add information that reflects the noise contamination level, which is $0 \leq s_i \leq 1, i$=$1,2,...,l$. Therefore, the training dataset $D$ becomes a triple $D^\prime$$=$$(\bm{x_i},y_i,s_i)$ to accommodate the fuzzy number and to reduce the influence of the contaminated data points in generating the decision functions. 
{\color{black}The fuzzy SVM is formulated as:}
\begin{equation} \label{eq:fuzzy}
	\begin{aligned}
		&\min_{\bm{\omega},b,\bm{\xi}}\;\; \frac{1}{2} ||\bm{\omega}||^2+C \bm{s^\top} \bm{\xi} \\
		&\;\; \text{s.t.  } y_i(\bm{\omega^\top} \bm{x_i}+b)+\xi_i \geq 1 \\
		&\;\; \xi_i \leq 0, i=1,2,...,l
	\end{aligned}
\end{equation}
\noindent where $C$ is the trade-off scalar and $\xi_i$ is the slack variable that represents the error associated with the $i$-th input data point. An important remark about this formulation is that a small $s_i$ can reduce the effect of the slack variable $\xi_i$ in Equation~\ref{eq:fuzzy}, so reducing the importance of the corresponding data point $\bm{x_i}$. The classification of an input $\bm{x}$ is given by the sign of $\bm{\omega^{*\top}} \bm{x}+b^*$, where $\bm{\omega^*}$ and $b^*$ are the solution of Equation~\ref{eq:fuzzy}. 

The construction of the membership functions follows the strategy used by Gao et al.~\cite{Gao2015,Gao2017}. The method considers reducing the noise carried by outliers while keeping the importance of the SVs. We integrate the fuzzy SVM into the TWSVM formulation by selecting two different classes and assigning a positive label to the first class and a negative label to the second one. The class centers $x_{c+}$ and $x_{c-}$ are the mean points considering the input space of these two classes, defined by:

\begin{equation}
	x_{c+}=\frac{1}{l+} \sum_{y_i=+1}\bm{x_i}, \;\; x_{c-}=\frac{1}{l-}\sum_{y_i=-1}\bm{x_i}
\end{equation}

The hyperspheres radii $r+$ and $r-$ are constructed by measuring the distance of the farthest scattering data point of each class:

\begin{equation}
	\begin{aligned}
		r_+=\max_i ||\bm{x_i}-x_{c+}||  \;\; \text{  if  }\;\;  y_i=+1 \\
		r_-=\max_i ||\bm{x_i}-x_{c-}||  \;\; \text{  if  }\;\;  y_i=-1 \\
	\end{aligned}
\end{equation}

\begin{figure}[h!]
	\centering
	\includegraphics[width=0.55\linewidth]{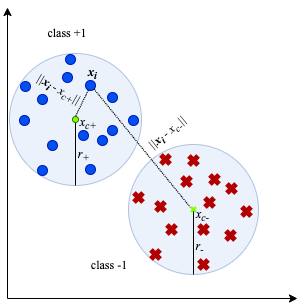}
	\caption{Graphical representation of $x_{c+}$, $x_{c-}$, $r_+$, $r_-$ and the distance between $||\bm{x_i}-x_{c+}||$ and $||\bm{x_i}-x_{c-}||$ given a positive instance $\bm{x_i}$.}
	\label{fig:Fuzzy_rep}
\end{figure}

The membership of $s_i$ is assigned according to the distance relationship between $||\bm{x_i}-x_{c+}||$ and $||\bm{x_i}-x_{c-}||$, when $x_{c+}$, $x_{c-}$, $r_+$, and $r_-$ are known. Formally, $s_i$ of a positive data point is given as:

\begin{equation}
	s_i=\begin{aligned} \begin{cases}
			\mu\big(1-||\bm{x_i}-x_{c+}||/(r_+ + \delta)  \big)
			\;\;\;\;\;\;\;\;\;\;\;\; \text{if }\;\;||\bm{x_i}-x_{c+}|| \geq ||\bm{x_i}-x_{c-}|| \; \land \; y_i=+1 \\
			(1-\mu)\big(1-||\bm{x_i}-x_{c+}||/(r_+ + \delta)  \big)
			\;\;\; \text{if }\;\;||\bm{x_i}-x_{c+}|| < ||\bm{x_i}-x_{c-}|| \;\land \; y_i=+1 
		\end{cases}
	\end{aligned}
\end{equation}

\noindent where $\mu\in [0,1]$ \textcolor{black}{balances} the effect of normal and noisy data points, and $\delta>0$ \textcolor{black}{avoids} fuzzy numbers equal 0. {\color{black} Figure~\ref{fig:Fuzzy_rep} illustrates the fuzzy-related elements used to assign the fuzzy  membership value.}

A data point is usually assigned by a proportional decreasing value $s_i$ when it drifts farther from its native class center, which increases the uncertainty \cite{Gao2017}. A small positive real number $\mu$ is assigned to decrease the effect of outliers towards the hyperplane. The fuzzy numbers for the negative data points are calculated \textcolor{black}{analogously}. 

\subsection{Kernel Approximation}
\label{sec:kernel}
Kernel machines that operate on the data kernel matrix (Gram matrix) scale more than quadratically in the data dimension \cite{Rahimi2008,Li2010}\textcolor{black}{, which} makes methods such as the ISVM or the LASVM impractical to deal with large datasets or incremental data that require sequential learning. Approximating non-linear kernels by linear kernels in the transformed space is a way to make possible the use of efficient linear methods that depend linearly on the size of the training set, allowing to solve large-scale and incremental learning problems efficiently \cite{Rahimi2008,Li2010}. Instead of relying on the kernel trick implicit lifting, the Random Fourier Features \cite{Rahimi2008} explicitly map the data to a low-dimensional Euclidean inner product using a randomized feature map $z\colon\mathcal{R}^n$$\rightarrow$$\mathcal{R}^N$, described as:
\begin{equation}
	\kappa(\bm{x_1},\bm{x_2})= \langle \varphi(\bm{x_1}),\varphi(\bm{x_2})\rangle  \approx z(\bm{x_1})^\top z(\bm{x_2})
\end{equation}
\noindent where $z$ is a low-dimensional space. The feature space approximates shift-invariant kernels $\kappa(\bm{x_1}-\bm{x_2})$ to within an error $err$ with $N$=$O( err^{-2}n \log\frac{1}{err^2})$ dimensions. Rahimi and Recth \cite{Rahimi2008} show empirically that a similar classification performance can be obtained for dimensions smaller than $N$.

The first set of transformed features are the Random Fourier bases $\cos (\tau^\top \bm{x}+b)$, where $\tau$$\in$$\mathcal{R}^n$ and $b$$\in$$\mathcal{R}$\textcolor{black}{, which are} random variables. It maps projected data on a randomly chosen line, followed by passing the resulting scalar through a sinusoidal function. The direction of these lines, in an appropriate distribution, guarantees that the product of two transformed points approximates a desired shift-invariant kernel \cite{Rahimi2008}. The transformation follows Bochner's theorem: \textit{A continuous kernel $\kappa(x,y)=\kappa(x-y)$ on $\mathcal{R}^n$ is positive definite if and only if $\kappa(\delta)$ is the Fourier transform of a non-negative measure.} For a properly scaled shift-invariant kernel $\kappa(\delta)$, Bochner's theorem guarantees that its Fourier transform $p(\tau)$ is a proper probability distribution:
\begin{equation}\label{eq:integral}
	\kappa(x-y)= \int_{\mathcal{R}^n} p(\tau) e^{j\tau^\top(x-y)}d\tau = E_\tau  [\zeta_\tau(x) \zeta_\tau(y)^*]
\end{equation} 
\noindent where $\zeta_\tau(x)=e^{j \tau^\top x}$. $\zeta_\tau(x)\zeta_\tau(y)^*$ is an unbiased estimate of $k(x,y)$ when $\tau$ is drawn for $p$, and $*$ denotes the complex conjugate. The integral of Equation~\ref{eq:integral} converges when the complex \textcolor{black}{exponential} are replaced by cosines, $z_\tau(x)$$=$$\sqrt{2}cos(\tau^\top x+b)$, obtaining a real-valued mapping that satisfies the condition $E[z_\tau(x) z_\tau(y)]$, where $\tau$ is drawn from $p(\tau)$ and $b$ is uniformly distributed from $[0, 2\pi]$. The \textcolor{black}{estimate kernel variance} can be reduced by concatenating $N$ randomly chosen $z_\tau$ into one $N$-dimensional normalized vector, i.e., the inner product $z(x)^\top z(y)$=$\frac{1}{N}\sum_{j=1}^{N}z_\tau(x) z_\tau(y)$ is a low variance approximation to the expectation of Equation~\ref{eq:integral}\footnote{The proof can be found in \cite{Rahimi2008}.}. To summarize, the random Fourier feature algorithm starts by getting a randomized feature map $z(x)$$:$$\mathcal{R}^n$$\rightarrow$$\mathcal{R}^N$, so that $z(x)^\top z(y)\approx k(x-y)$. The second step is to compute \textcolor{black}{$p$ of the kernel, which in this case is the Fourier transform of $k$:}

\begin{equation}\label{eq:Fouriertransform}
	p(\tau)= \frac{1}{2\pi} \int e^{j\tau^\top\delta} k(\delta)d \Delta
\end{equation}

The third step is to draw $N$ independent and identically distributed (iid) data points $\tau_1,...,\tau_N\in\mathcal{R}^n$ from $p$ and $N$ iid data points $b_1,...,b_N\in\mathcal{R}$ from the uniform distribution on $[0, 2\pi]$. Finally, $z(\bm{x})$ is computed as:

\begin{equation}
	z(\bm{x})\equiv \sqrt{\frac{2}{N}}[cos(\tau_1^\top \bm{x}+b_1),\dots, cos(\tau_N^\top\bm{x}+b_N)]^\top
\end{equation}

The scalar $\sigma^2_p$ is equal to the trace of {\color{black}the} Hessian of $k$ at 0, that quantifies the curvature of the kernel at the origin. For a Gaussian kernel denoted as $k(\bm{x_1},\bm{x_2})$$=$$\exp(-\gamma||\bm{x_1}-\bm{x_2}||^2)$, we have $\sigma^2_p$$=$$2n\gamma$, that approximates the kernel to:

\begin{equation}
	p(\bm{\tau})=2\pi^{-\frac{N}{2}}\exp(-\frac{||\tau||^2_2}{2})
\end{equation}

The important implications of using this kernel approximation in our incremental approach are: (i) we approximate the non-linear model accuracy with a linear model; (ii) it is faster to calculate the approximate kernel than the regular one; (iii) and mainly, we increase the model only in one dimension, so we do not need to recalculate the kernel approximation of the previous data. 

\section{The Fuzzy Bounded Twin SVM (FBTWSVM)}
\label{se:FBTSVM}

We propose a formulation based on the TWSVM \cite{Jayadeva2007} and inspired by the FRTSVM \cite{Gao2015,Gao2017} and TBSVM \cite{Yuan-HaiShao2011} properties. The fuzzy formulation incorporated by our method (Equation~\ref{eq:fuzzy}) is inspired by the TWSVM (Equations~\ref{eq:TWSVM+} and~\ref{eq:TWSVM-}), while the adherence to the structural risk minimization principle is incorporated similarly to the TBSVM \cite{Yuan-HaiShao2011}. To maintain such an adherence, we need to guarantee the existence of the dual formulation inverse matrix, which circumvents the drawback of the standard TWSVM (i.e., the standard TWSVM only adheres to the empirical risk minimization problem in the dual problem). Thus, we define the FBTWSVM primal formulation as:
\begin{equation}\label{eq:FRBTSVM1}
	\begin{split}
		& \min_{\bm{\omega_+},b_+,\bm{\xi_-}} \;\; \frac{1}{2} C_1 (||\bm{\omega_+}||^2+b^2_+)+\frac{1}{2}||X_+ \;\bm{\omega_+} + \bm{e_+} \; b_+||^2 + C_3 \bm{s^\top_-} \; \bm{\xi_-} \\
		&\;\;\text{s.t. } y_-(X_- \; \bm{\omega_+} + \bm{e_-} \; b_+)+ \bm{\xi_-} \geq \bm{e_-}, \;\; \bm{\xi_-} \geq 0
	\end{split}
\end{equation}
\vspace{-20pt}
\begin{equation}\label{eq:FRBTSVM2}
	\begin{split}
		&\min_{\bm{\omega_-},b_-,\bm{\xi_+}}\;\; \frac{1}{2} C_2 (||\bm{\omega_-}||^2+b^2_-)+\frac{1}{2}||X_- \; \bm{\omega_-} + \bm{e_-} \; b_-||^2 + C_4 \bm{s^\top_+} \; \bm{\xi_+} \\
		&\;\;\text{s.t. }y_+(X_+ \;\bm{\omega_-} + \bm{e_+} \; b_-)+ \bm{\xi_+} \geq \bm{e_+}, \;\; \bm{\xi_+} \geq 0
	\end{split}
\end{equation}

\noindent where $C_1$, $C_2$, $C_3$, and $C_4$ are the trade-off parameters between the margin and the complexity for weighting the regularization, $\bm{s_+}$$\in$$R^{l_+}$ and $\bm{s_-}$$\in$$R^{l_-}$ are the fuzzy number vectors sequentially associated with the positive and negative input data points, which introduce the desired robustness in the weighted regularized model \cite{Gao2015,Gao2017}. The additional $b_+$ and $b_-$ in Equations~\ref{eq:FRBTSVM1} and~\ref{eq:FRBTSVM2} minimize the structural risk.

The two hyperplanes in $\mathcal{R}^n$ are defined as $\bm{\omega^\top_\pm}+b_\pm$=$0$, and since the TWSVM has two proximal decision functions, two margin terms $1/||\bm{\omega_\pm}||$ are defined for the proximal decision function \cite{Gao2015}. The margin between two classes can be measured by the distance between the proximal hyperplane $\bm{x^\top} \bm{\omega_+}+b_+$=$0$ and the bounding hyperplane $\bm{x^\top}\bm{\omega_+}+b_+$=$-1$. The distance is $1/||\bm{\omega_+}||^2$, and it is the one-sided margin between the two classes with respect to the hyperplane $\bm{x^\top}\bm{\omega_+}+b_+$=$0$ \cite{Yuan-HaiShao2011,Jayadeva2017}. The process is analogous to the other hyperplane. We need to write the dual problems {\color{black} obtaining} the solutions of Equations~\ref{eq:FRBTSVM1} and~\ref{eq:FRBTSVM2}. We start by taking the Lagrangian of Equation~\ref{eq:FRBTSVM1} to obtain the Wolfe dual:
\begin{equation}
	\begin{split}
		& L(\bm{\omega_+},b_+,\bm{\xi_-})=\frac{1}{2}C_1 (||\bm{\omega_+}||^2+b^2_+) +\frac{1}{2}||X_+\bm{\omega_+}+\bm{e_+} b_+||^2\\ 
		& \qquad\qquad -\bm{\alpha^\top}(-(X_- \; \bm{\omega_+}+\bm{e_-} \; b_+)+\xi_- \;-\;\bm{e_-})+C_3 \; \bm{s^\top_-} \; \bm{\xi_-}-\eta^\top\bm{\xi_-}
	\end{split}
\end{equation}

\noindent where $\bm{\alpha}$$=$$(\alpha_1,\dots,\alpha_{X_+})^\top$, and $\bm{\eta}$$=$$(\eta_1,\dots,\eta_{X_+})^\top$ are the Lagrange multiplier vectors. Considering that Equation~\ref{eq:FRBTSVM1} {\color{black}represents} a convex optimization problem, the Karush-Kuhn-Tucker (KKT) optimality conditions are both necessary and sufficient, and they are written as:
\begin{subequations}
	\begin{align}
		& \nabla \bm{\omega_+} \; L=C_1 \; \bm{\omega_+} + X^\top_+ (X_+ \; \bm{\omega_+} + \bm{e_+} \; b_+) + X^\top_- \; \bm{\alpha}=0
		\label{eq:KKT1}\\        
		& \nabla b_+ \; L=C_1 \; b_+ + \bm{e^\top_+} (X_+ \; \bm{\omega_+} + \bm{e_+} \; b_+)+\bm{e^\top_-} \; \bm{\alpha}=0 
		\label{eq:KKT2}\\
		& \nabla \bm{\xi_-} \; L= -\bm{\alpha^\top} - \bm{\eta^\top}+C_3 \; \bm{s_-}=0 \label{eq:KKT3}\\
		& -(X_- \; \bm{\omega_+} + \bm{e_-} \; b_+)+\bm{\xi_-}\;\geq\;\bm{e_-}\bm{\xi_-}\;\geq 0 \label{eq:KKT4}\\
		& \bm{\alpha^\top}(\bm{\omega_-} \; X_+ + \bm{e_-} \; b_+ -\bm{\xi_-} + \bm{e_-})=0; \;\;\; \bm{\eta^\top} \bm{\xi_-}=0 \label{eq:KKT5} \\
		& \bm{\alpha} \geq 0, \;\;\bm{\eta} \geq 0, \;\;\; \bm{\xi_-} \geq 0 \label{eq:KKT6}
	\end{align}
\end{subequations}

Considering that $\eta\geq 0$ and $\alpha \geq 0$ from Equation~\ref{eq:KKT6}, and using Equation~\ref{eq:KKT3}, we know that $\alpha$ is bounded as $0 \leq \alpha \leq C_3 s_-$. Summing Equations~\ref{eq:KKT1} and~\ref{eq:KKT2}, and using Equations~\ref{eq:KKT3} to ~\ref{eq:KKT6} for simplification, we obtain:
\begin{equation}\label{eq:combKKT}
	([X_+,\bm{e_+}]^\top [X_+,\bm{e_+}]+C_1 I )[\bm{\omega_+},b_+]+ [X_-,\bm{e_-}]^\top \bm{\alpha}=0
\end{equation}

Defining $H_+$$=$$[X_+,\bm{e_+}]$, $H_-$$=$$[X_-,\bm{e_-}]$, $\bm{u_+}$$=$$[\bm{\omega_+},b_+]$ and $\bm{u_-}$$=$$[\omega_-,b_-]$ (one to each class), we can rewrite Equation~\ref{eq:combKKT} as:
\begin{equation}\label{eq:comb2kkt}
	\begin{aligned}
		&(H^\top_+ H_+ + C_1 I)\bm{u_+^\top} + H^\top_- \; \bm{\alpha} =0 \;\;\;\; \text{or}\\
		&\bm{u_+^\top}= - (H^\top_+ H_+ + C_1 I)^{-1}H^\top_- \; \bm{\alpha}
	\end{aligned}
\end{equation}

Using our notation, the Wolfe dual is defined as:
\begin{equation}
	\begin{aligned}
		&\max \;\; L(\bm{\omega_+},b_+,\bm{\xi_-}, \bm{\alpha}, \bm{\eta}) \\
		& \;\; \text{s.t } \;\;\; \nabla_{\bm{\omega_+}} L(\bm{\omega_+},b_+,\bm{\xi_-}, \bm{\alpha}, \bm{\eta}) \\
		&\quad \quad  \frac{\partial L}{\partial b_+}=0 \\
		&\quad \quad  \frac{\partial L}{\partial \bm{\xi_-}}=0\\
		&\quad \quad  \bm{\alpha} \geq 0,\;\; \bm{\eta} \geq 0
	\end{aligned}
\end{equation}

Using the KKT conditions (from Equations~\ref{eq:KKT1} to~\ref{eq:KKT6}) and Equation~\ref{eq:comb2kkt}, the Wolfe dual of Equations~\ref{eq:FRBTSVM1} and~\ref{eq:FRBTSVM2} can be written as:

\begin{equation}\label{eq:dualFRBTSVM1}
	\begin{aligned}
		&\max_{\bm{\alpha}} \;\; \bm{e^\top_-} \; \bm{\alpha} -\frac{1}{2} \bm{\alpha^\top} H_- (H^\top_+ H_+ + C_1 I_1)^{-1}H^\top_- \; \bm{\alpha} \\
		& \;\; \text{s.t. } 0 \leq \bm{\alpha} \leq C_3 \bm{s_-}
	\end{aligned}
\end{equation}
\begin{equation}\label{eq:dualFRBTSVM2}
	\begin{aligned}
		&\max_{\bm{\nu}} \;\; \bm{e^\top_+} \; \bm{\nu} -\frac{1}{2} \bm{\nu^\top} H_+ (H^\top_- H_- + C_2 I_2)^{-1}H^\top_+ \; \bm{\nu} \\
		& \;\;  \text{s.t. } 0 \leq \bm{\nu} \leq C_4 \bm{s_+}
	\end{aligned}
\end{equation}

\noindent where $I_1$ and $I_2$ are identity matrices. The matrices $(H^\top_+ H_+ + C_1 I_1)$  and $(H^\top_- H_- + C_2 I_2)$ from Equations~\ref{eq:dualFRBTSVM1} and~\ref{eq:dualFRBTSVM2} are non-singular naturally, therefore their inverses are guaranteed to exist, which adds the adherence to the structural risk minimization principle \cite{Yuan-HaiShao2011,Jayadeva2017}. Notice that the dual for Equation~\ref{eq:FRBTSVM2} can be obtained is an analogous way. By solving the duals (Equations~\ref{eq:dualFRBTSVM1} and~\ref{eq:dualFRBTSVM2}), we obtain the optimal solutions for $\bm{\alpha^*}$ and $\bm{\nu^*}$, and furthermore, the corresponding classes $\bm{u^*_\pm}$ (as defined in Equation~\ref{eq:comb2kkt}) and the non-parallel hyperplanes. The dual of Equation~\ref{eq:dualFRBTSVM1} and~\ref{eq:dualFRBTSVM2} relates to the primal problems (Equations~\ref{eq:FRBTSVM1} and~\ref{eq:FRBTSVM2}) as:

\begin{equation}
	\begin{aligned}
		\bm{u^*_+} = -(H^\top_+H_+ + C_1 I_1)^{-1} H^\top_- \bm{\alpha^*} \\
		\bm{u^*_-} = (H^\top_-H_- + C_2 I_2)^{-1} H^\top_+ \bm{\nu^*}
	\end{aligned}
\end{equation}

Finally, for a test data point $\bm{x}\in\mathcal{R}^n$, the classification decision function is given by Equation~\ref{eq:Classif}.

\subsection{The Non-linear FBTWSVM}
In the non-linear FBTWSVM, the input data points $x\in\mathcal{R}^n$ are mapped to a high-dimensional space $\mathcal{H}$ through $\varphi(x)$. The kernel function $\kappa(\cdot,\cdot)$ calculates implicitly the dot product of a pair of transformations, which is applied as $\kappa(x_1,x_2)=\langle \varphi(x_1),\varphi(x_2)\rangle $. The non-linear dual proximal hyperplanes are:

\begin{equation}
	\label{eq:nonlinearsurfaces}
	\begin{aligned}
		\kappa(x,x^\top)\bm{\omega_+}+b_+=0 \\
		\kappa(x,x^\top)\bm{\omega_-}+b_-=0
	\end{aligned}
\end{equation}
and the primal problems used to obtain the dual proximal hyperplanes are:

\begin{equation}
	\label{eq:nonFRBTSVM1}
	\begin{split}
		& \min_{\bm{\omega_+},b_+,\bm{\xi_-}} \;\; \frac{1}{2} C_1 (||\bm{\omega_+}||^2 + b^2_+) +  C_3 \; \bm{s^\top_-} \; \bm{\xi_-} + \frac{1}{2} || \kappa(X_+,X^\top)\bm{\omega_+} + \bm{e_+} \; b_+||^2  \\
		&\;\;\;\; \text{s.t. } y_-(\kappa(X_-,X^\top)\bm{\omega_+} + \bm{e_-} \; b_+ ) + \bm{\xi_-} \geq \bm{e_-}, \;\; \bm{\xi_-} \geq 0
	\end{split}
\end{equation}
\vspace{-20pt}
\begin{equation}
	\label{eq:nonFRBTSVM2}
	\begin{split}
		& \min_{\bm{\omega_-},b_-,\bm{\xi_+}} \;\; \frac{1}{2} C_2 (||\bm{\omega_-}||^2 + b^2_-) + C_4 \; \bm{s^\top_+} \; \bm{\xi_+}+\frac{1}{2} || \kappa(X_+,X^\top)\bm{\omega_-} + \bm{e_-} \; b_-||^2 \\
		&\;\;\;\; \text{s.t. } y_+(\kappa(X_+,X^\top)\bm{\omega_-} + \bm{e_+} \; b_- ) + \bm{\xi_+} \geq \bm{e_+}, \;\; \bm{\xi_+} \geq 0
	\end{split}
\end{equation}

The dual forms of Equation~\ref{eq:nonFRBTSVM1} and~\ref{eq:nonFRBTSVM2} are:
\begin{equation}\label{eq:dualnonFRBTSVM1}
	\begin{aligned}
		&\max_{\bm{\alpha}} \;\; \bm{e^\top_-} \; \bm{\alpha} -\frac{1}{2} \bm{\alpha^\top} S_- (S^\top_+ S_+ + C_1 I_1)^{-1}S^\top_- \; \bm{\alpha} \\
		& \;\; \text{s.t. } 0 \leq \bm{\alpha} \leq C_3 \bm{s_-}
	\end{aligned}
\end{equation}
\vspace{-10pt}
\begin{equation}\label{eq:dualnonFRBTSVM2}
	\begin{aligned}
		&\max_{\bm{\nu}} \;\; \bm{e^\top_+} \; \bm{\nu} -\frac{1}{2} \bm{\nu^\top} S_+ (S^\top_- S_- + C_2 I_2)^{-1}S^\top_+ \; \bm{\nu} \\
		& \;\;  \text{s.t. } 0 \leq \bm{\nu} \leq C_4 \bm{s_+}
	\end{aligned}
\end{equation}

\noindent where $S_+=[\kappa(X_+,X^\top),\bm{e_+}]$ and $S_-=[\kappa(X_-,X^\top),\bm{e_-}]$. The solutions of the primal problems of Equations~\ref{eq:nonFRBTSVM1} and~\ref{eq:nonFRBTSVM2} are $\bm{\upsilon^*_\pm}=[\bm{\omega^{*\top}_\pm}, b^*_\pm]^\top$, which are the parametric relationships between the optimal $\bm{\upsilon^*_\pm}$ and the optimal solutions $\bm{\alpha^*}$ and $\bm{\nu^*}$ of the dual forms of Equations~\ref{eq:dualnonFRBTSVM1} and~\ref{eq:dualnonFRBTSVM2}:

\begin{equation}
	\begin{aligned}
		& \bm{\upsilon^*_+} = -(S^\top_+ S_+ + C_1 I_1)^{-1}S^\top_- \; \alpha^* \\
		& \bm{\upsilon^*_-} = (S^\top_- S_- + C_2 I_2)^{-1}S^\top_+ \; \nu^* \\
	\end{aligned}
\end{equation}

Once Equations~\ref{eq:dualnonFRBTSVM1} and~\ref{eq:dualnonFRBTSVM2} are solved to obtain the hyperplanes (Equation~\ref{eq:nonlinearsurfaces}), a new data point $\bm{x}\in\mathcal{R}^n$ can be classified in a similar manner to the linear case by Equation~\ref{eq:Classif}.

\subsection{Solving The FBTWSVM}
\textcolor{black}{We use the} coordinate descent method (DCD) \cite{Chang2008} \textcolor{black}{to solve the dual problem of the FBTWSVM \cite{Shao2012}}. The DCD leads to fast training by updating one variable at a time through a single-variable sub-problem minimization. Such fast training allows the processing of large and incremental datasets \cite{Shao2012}. The dual problems of Equations~\ref{eq:dualFRBTSVM1} and~\ref{eq:dualFRBTSVM2} and Equations~\ref{eq:dualnonFRBTSVM1} and~\ref{eq:dualnonFRBTSVM2} are solved in the same way. However, for convenience, we only present the solution of Equation~\ref{eq:dualFRBTSVM1}. We start by considering $Q$$=$$H_-(H^\top_+H_++C_1 I_1)^{-1}H_-^\top$ and $Q^\prime$$=$$(H^\top_+H_++C_1 I_1)^{-1}H_-^\top$. Consequently, $Q$$=$$H_-Q^\prime$, where $q_{ii}$ and $\overline{Q}$ can be pre-computed and stored if necessary. The matrix inversion is calculated with the Sherman-Morison-Woodbury formula. Assuming $\bm{\alpha^{k,i}}$$=$$[\alpha^{k+1,i}_1,\dots,\alpha^{k+1,i}_{i-1},\alpha^{k,i}_i,\dots,\alpha^{k,1}_{X_-+1}]$, where $i$$=$$(1,\dots,X_-$$+$$1)$ is the index for the data points and $k=(-1,+1)$ is the data label. We use the following problem updating from $\bm{\alpha^{k,i}}$ to $\bm{\alpha^{k,i+1}}$:

\begin{equation}\label{eq:DCD1}
	\begin{aligned}
		\min_{d} \;\; f(\bm{\alpha^{k,i}}+d \; \bm{e_i}) \\
		\;\; \text{s.t. } 0\leq\bm{\alpha^k_i}+d\leq C_3\bm{s_-}
	\end{aligned}
\end{equation}

\noindent where $\bm{e_i}$$=$$[0,\dots,0,1,0,\dots,0]^\top$ (the $i-$th position is 1), and $\bm{d_i}$ is an optimum solution to the problem of minimizing $f(\bm{\alpha^{k,i}}+d\bm{e_i})$ subject to $\bm{d_i}\in\mathcal{R}^n$, i.e., $f(\bm{\alpha^{k,i}}+d\bm{e_i})$ achieves a minimum at $\bm{d_i}$ only if $\nabla f(\bm{\alpha^{k,i}}+d\bm{e_i})^\top \bm{e_i}=\nabla f(\bm{\alpha^{k,i+1}})^\top\bm{e_i} = 0$\footnote{The proof can be found in \cite{Chang2008}}. The objective function of Equation~\ref{eq:DCD1} is a quadratic function of $d$:

\begin{equation}\label{eq:DCDcte}
	f(\bm{\alpha^{k,i}}+d\bm{e_i})=\frac{1}{2}Q_{ii}\bm{d^2}+\nabla_if(\bm{\alpha^{k,i}})d+\text{constant}
\end{equation}

\noindent where $\nabla_if$ is the $i$-th component of the gradient $\nabla f$. Equation~\ref{eq:DCD1} has an optimum at $d$=$0$ iff:

\begin{equation}\label{eq:DCD2}
	\nabla^P_if(\bm{\alpha^{k,i}})=0
\end{equation}

\noindent where $\nabla^P_if(\alpha)$ is the projected gradient which is defined as:

\begin{equation}\label{eq:projected_grad}
	\nabla^P_if(\alpha)=\begin{aligned}\begin{cases}
			\min(0,\nabla_if(\alpha)), \quad \alpha_i=0, \\
			\nabla_if(\alpha), \quad \quad \quad \quad \;\;  0\leq \alpha_i \leq C_3s_-\\
			\max(0,\nabla_if(\alpha)), \quad \alpha_i=C_3 s_-
		\end{cases}
	\end{aligned}
\end{equation}

If Equation~\ref{eq:DCD2} is satisfied, we can move to the next iteration ($i$+1) without updating $\bm{\alpha^{k,i}_i}$ in $X_-$, i.e., we only update $\bm{\alpha^{k,i}_i}$ to temporally meet the optimal solution of Equation~\ref{eq:DCD1}. The optimum of Equation~\ref{eq:DCDcte} is reached by introducing the Lipschitz continuity:

\begin{equation}\label{eq:DCDminmax1}
	\bm{\alpha^{k,i+1}_i} =\min(\max(\bm{\alpha^{k,i}_i}-\nabla_if(\bm{\alpha^{k,i}})/Q^\prime_{ii},0),C_3s_{i_-})
\end{equation}

In the update of Equation~\ref{eq:DCDminmax1}, $Q^\prime_{i,i}$ can be pre-calculated by $Q^\prime_{ii}=H_{-i}Q_i$, and $\nabla_if(\bm{\alpha^{k,i}})$ can be obtained by:

\begin{equation}\label{eq:DCD4}
	\nabla_if(\alpha)=(Q'\alpha)_i-1=\sum_{j=1}^{X_-}Q^\prime_{ij}\alpha_j-1
\end{equation}

The computation of Equation~\ref{eq:DCD4} is approximated as $O(X_-\overline{l})$, where $\overline{l}$ is the average count of non-zero elements in $Q^\prime$ per data point. To reduce the number of operations, we can alternatively compute Equation~\ref{eq:DCD4} as:

\begin{equation}\label{eq:u_plus}
	\nabla_if(\alpha)=-H_{-i}u_+-1
\end{equation}

\noindent with a pre-defined $\bm{u_+}$$=$$-Q\alpha$ and $i$ is the row of the matrix $H_-$, so the number of operations is $O(\overline{n})$. To maintain $\bm{u_+}$ throughout {\color{black}the} coordinate descent procedure, we use:

\begin{equation}\label{eq:DCDite}
	\bm{u_{+i}} \leftarrow \bm{u_{+i}} - Q_i (\alpha_i - \overline{\alpha}_i)
\end{equation}

The complexity to maintain $\bm{u_+}$ iteratively is $O(\overline{l})$. Starting with $\alpha^0=0$, the optimal solution of $\bm{u_+}$ is obtained by iterative updating Equation~\ref{eq:DCDite}, and furthermore, the optimal solution of Equation~\ref{eq:dualFRBTSVM1}. The cost per iteration for the whole process is $O(X_2 \overline{n})$, and the memory requirement is the size of $H_-$ and $Q^\prime$.

\subsection{Implementation}
The dual problem of Equation~\ref{eq:dualFRBTSVM1} has the constraint $0\leq\alpha_i\leq C_3\bm{s_-}$, and if $\alpha_i$ is either $0$ or $C_3\bm{s_-}$, it may achieve a steady state. Considering that our formulation produces many bounded Lagrange multipliers, we apply the proposed shrinking technique to reduce the size of the optimization problem without considering some bounded variables \cite{Joachims1999}. Considering $Z$ as a subset of $X$ after removing all data points that have non-bounded Lagrange multipliers, and $\overline{Z}$$=$$\{1,\dots,X_- \}/Z$ its complement subset, the dual of Equation~\ref{eq:dualFRBTSVM1} can be represented by a smaller problem that consumes less time and memory:

\begin{equation}\label{eq:imp1}
	\begin{aligned}
		& \min_{\bm{\alpha}_Z}\frac{1}{2}\bm{\alpha}^\top_ZQ'_{ZZ}\bm{\alpha}_Z+(Q^\prime_{Z\overline{Z}}\bm{\alpha}_{\overline{Z}}-\bm{e}_Z)^\top\bm{\alpha}_Z \\
		& \quad\text{s.t. }0\leq\bm{\alpha}_Z\leq C_3\bm{s_}{-Z}
	\end{aligned}
\end{equation}

\noindent where $Q_{ZZ}$ and $Q_{Z\overline{Z}}$ are sub-matrices of $Z$ and $\bm{\alpha}_{\overline{Z}}$ is a vector of Lagrangian multipliers. To solve Equation~\ref{eq:imp1}, we compute $\nabla_if(\bm{\alpha})$ as:

\begin{equation}
	\nabla_if(\bm{\alpha})=Q_{i,Z}\bm{\alpha}_Z+Q_{i,\overline{Z}}\bm{\alpha}_{\overline{Z}}-1
\end{equation}
\noindent If $i\in Z$, and defining $\bm{u_1}$ as:
\begin{equation}
	\bm{u_1}=-(Q^\prime_{i\in Z}\bm{\alpha}_{i\in Z}+Q^\prime_{i\in\overline{Z}}\bm{\alpha}_{i\in\overline{Z}})
\end{equation}

\noindent \textcolor{black}{we} have $\nabla_if(\bm{\alpha})$=$H_-\bm{u_1}-$1, which turns $\nabla_if(\alpha)$ easy to obtain. For a linear kernel we only need to update ($\overline{Q}_{i\in Z}\bm{\alpha}_{i\in Z}$), and we do not need to reconstruct all $\nabla f(\bm{\alpha})$ to implement the shrink step\footnote{The proof can be found in \cite{Hsieh2008}.}. Considering the projected gradient $\nabla^Pf(\bm{\alpha})$ defined in Equation~\ref{eq:projected_grad}, and following the optimality condition of bound-constrained problems, $\alpha$ is optimal iff $\nabla^P f(\bm{\alpha})=0$. During the iteration procedure, the inequality $\nabla^Pf(\bm{\alpha})\neq 0$ means either $\max_j\nabla^P f(\bm{\alpha})>0$ or $\min_j\nabla^Pf(\bm{\alpha})<0$, and at the $k-$1 step, we obtain $m_{\max}^{k-1}\equiv\max_j\nabla^P f(\bm{\alpha})$ and $m_{\min}^{k-1}\equiv\min_j\nabla^P f(\bm{\alpha})$. In this way, at each inner step of the $k-$th iteration, and before updating $\bm{\alpha}$, the element is shrunken if one of the two conditions holds:

\begin{equation}\label{eq:DCDshrink}
	\begin{aligned}
		&\alpha^{k,i}_i=0 \quad \text{and} \quad \nabla^P f(\bm{\alpha^{k,i}})> {m^\prime}_{\max}^{k-1} \\
		\text{or}\\
		&\alpha^{k,i}_i=C_3 \bm{s_-} \quad \text{and} \quad \nabla^P f(\bm{\alpha^{k,i}})< {m^\prime}_{\min}^{k-1}
	\end{aligned}
\end{equation}

\noindent where ${m^\prime}_{\max}^{k-1}$ must be strictly positive and ${m^\prime}_{\min}^{k-1}$ must be strictly negative, and they are defined as:

\begin{equation}
	\begin{aligned}
		{m^\prime}_{\max}^{k-1}=\begin{cases}
			m_{\max}^{k-1}, \quad \text{if }m_{\max}^{k-1}>0,\\
			\infty, \quad \quad \quad  otherwise
		\end{cases}
	\end{aligned}
\end{equation}
\begin{equation}
	\begin{aligned}
		{m^\prime}_{\min}^{k-1}=\begin{cases}
			m_{\max}^{k-1}, \quad \text{if }m_{\min}^{k-1}<0,\\
			-\infty, \quad \quad \quad  otherwise
		\end{cases}
	\end{aligned}
\end{equation}

Next, we multiply both ${m^\prime}_{\max}^k-$$1$ and ${m^\prime}_{\min}^k-$$1$ by a shrinking rate smaller than one. A tolerance $\epsilon$ indicates if the optimal value is satisfied after a finite number of iterations, thus it is used as a valid stop criterion:
\begin{equation}\label{eq:imp2}
	{m'}_{\max}^k - {m'}_{\min}^k < \epsilon
\end{equation}

If in the $k-$th iteration, the condition stated in Equation~\ref{eq:imp2} is satisfied for Equation~\ref{eq:imp1}, we can enlarge the active set $Z$ to $\{1,\dots,X_-+$1$\}$, and set ${m^\prime}_{\max}^k=+\infty$ and ${m^\prime}_{\min}^k=-\infty$, and continue with the regular iterations. We store the previous values of ${m^\prime}_{\max}$ and ${m^\prime}_{\min}$ during the DCD process to avoid recalculating them during the incremental step. Therefore, the shrinking technique is a key step to avoid calculating and storing all training data during the training phase. Our method process one class at each time, however, the inner processing can be done in parallel, where one input is assigned to an available processor to calculate the membership followed by the Lagrangian multiplier. We present the pseudo-code of the FBTWSVM training algorithm (Algorithm~\ref{alg:Training}) for the positive class.

\subsection{The Multiclass FBTWSVM}
\label{se:multiclass}
The FBTWSVM is \textcolor{black}{based on} the TWSVM \textcolor{black}{foundations}, which considers only binary problems. \textcolor{black}{Yet}, we can extend the FBTWSVM to multiple classes by building and combining several binary classifiers instead of considering all data in one optimization formula \cite{Chih-WeiHsu2002}. The multiclass FBTWSVM is based on the Decision Directed Acyclic Graph (DDAG), which achieves better accuracy while requiring less training time than other multiclass approaches \cite{Tomar2015,Platt2000a}. The DAG-based multiclass classifier was originally proposed by Platt et al.~\cite{Platt2000a} for the multiclass SVM approach, and further introduced by Chen and Ji \cite{JingChen2010} into the twin approach as the Optimal DAG to the Least Squares Twin SVM.

\begin{figure}[htpb!]
	\centering
	\includegraphics[width=0.7\linewidth]{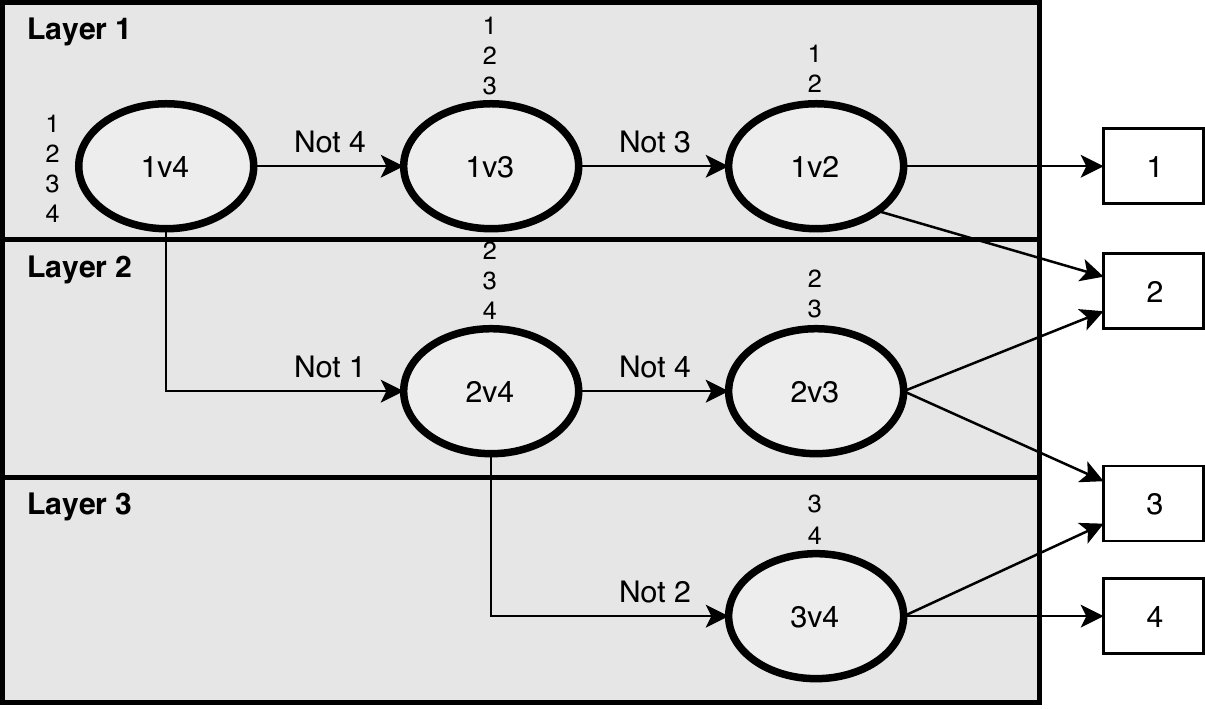}
	\caption{A 4-class classification problem based one the DAG topology.}
	\label{fig:DAG}
\end{figure}

In the multiclass approach based on the DAG topology, for a $u-$class classification problem, there are $u(u-$$1)/2$ sub-classifier nodes divided into $u-$$1$ layers. During the classification process, there is no need for combining all sub-classifiers, so to assign a class to a test data point, it makes $u$$-$$1$ decisions. The classification process starts at the root node, located in the first layer, and includes all possible classification labels (node 1v4 in Figure~\ref{fig:DAG}). The decision-making step eliminates the most excluded category at each sub-classifier decision, i.e., considering a 4-class problem with a test data point with label $y_i$$=$$4$ and the topology presented in Figure~\ref{fig:DAG}. The root node sub-classifier eliminates the possibility of $y_i$$=$$1$, following the \textit{Not} 1 line. The next sub-classifier eliminates the possibility of $y_i$$=$$2$ following the \textit{Not} 2, and the last sub-classifier eliminates the possibility of $y_i$$=$$3$, assigning class 4 to the test data point.

\begin{algorithm}[!tb]\footnotesize
	\caption{FBTWSVM's training procedure}\label{alg:Training}\footnotesize
	\textbf{Input:} X, \textbf{y}, $C_1$,$C_2$,$C_3$,$C_4$\\
	\textbf{Output:} $\bm{u_+}$
	\begin{algorithmic}[1]
		\State Compute $Q=(H^\top_{+} H_++C_1 I)^{-1}H^\top_-$ and ${Q^\prime}_{ii}=H_{-i}Q_i$
		\State Let $Z=\{1,\dots,X_- \}$
		\State Given $\epsilon$, $\bm{\alpha}=0$ and $\bm{u_+}=0$
		\State ${m^\prime}_{\max}=\infty$ and ${m^\prime}_{\min}=-\infty$
		\While $ $
		\State Let $m_{\max}=- \infty$, $m_{\min}=\infty$
		\For {$\forall i \in Z$, (a randomly and exclusively selected)} \Comment{this thread runs in parallel}
		\State $\nabla_i \; f(\bm{\alpha})=-H_{-i}\bm{u_+}-1$
		\State $\nabla^P_i \; f(\bm{\alpha})=0$
		\If {$\alpha_i=0$}
		\If {$\nabla^P_i \; f(\bm{\alpha})>{m^\prime_{\max}}$} $X=X/ \{i\}$ \EndIf
		\If {$\nabla^P_i \; f(\bm{\alpha})<0$} $\nabla^P_i \; f(\bm{\alpha}) = \nabla_if(\bm{\alpha})$ \EndIf
		\If {$\alpha_i=C_3 s_-$}
		\If {$\nabla^P_i \; f(\bm{\alpha})<{m^\prime}_{\min}$} $X=X/ \{i\}$ \EndIf
		\If {$\nabla^P_i \; f(\bm{\alpha})>0$} $\nabla^P_i \; f(\bm{\alpha}) = \nabla_if(\bm{\alpha})$ \EndIf
		\EndIf
		\Else
		$ \nabla^P_if(\bm{\alpha})=\nabla_if(\bm{\alpha})$
		\EndIf
		\State $m_{\max}=\max(m_{\max},\nabla^P_if(\bm{\alpha}))$
		\State $m_{\min}=\min(m_{\min},\nabla^P_if(\bm{\alpha}))$
		\If {$\nabla^P_if(\bm{\alpha})\neq 0$}
		\State $\overline{\alpha}=\alpha_i$
		\State $\alpha_i=\min(\max(\alpha_i-\nabla_if(\bm{\alpha})/{Q^\prime}_{ii},0 )C_3s_{i-})$
		\State $u_{+i}=u_{+i}-Q_i(\alpha_i-\overline{\alpha_i})$
		\EndIf
		\EndFor
		\If {$m_{\max}-m_{\min}<\epsilon$}
		\If {$X=\{1,\dots,l_- \}$,} \textbf{break} \EndIf
		\Else 
		\State $X=\{1,\dots,l_- \}, {m^\prime}_{\max}=\infty, {m^\prime}_{\min}=-\infty$
		\EndIf
		\If {$m_{\max} \leq 0$} ${m^\prime}_{\max}=\infty$ \Else $\;\; {m^\prime}_{\max}=m_{\max}$ \EndIf
		\If {$m_{\max} \geq 0$} ${m^\prime}_{\min}=-\infty$ \Else $\;\; {m^\prime}_{\min}=m_{\min}$ \EndIf
		\EndWhile
		\State\Return $\bm{u_+}$
	\end{algorithmic}
\end{algorithm}

\section{Incremental and Decremental FBTWSVM}
\label{sec:inc_dec}
The FBTWSVM can integrate new data points continuously into the existing model without fully reconstructing it.  Besides that, it can be trained fast due to the formulation and the solver choice, and it generalizes well like a conventional SVM. These characteristics make it suitable for incremental learning applications. 

\begin{figure}[tb!]
	\centering
	\includegraphics[width=0.55\linewidth]{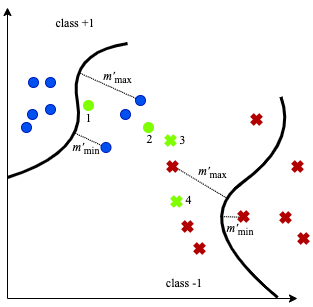}
	\caption{Blue circles and red crosses represent classes respectively, and the green circles and crosses represent new data points from each class.}
	\label{fig:Inc_FBTSVM}
\end{figure}

The incremental FBTWSVM is based on the shrinking heuristic that can increase the current model considering the fuzzy information of new values. We update the model by selecting only new data points that extrapolate the minimum (${m^\prime}_{\min}$) and maximum (${m^\prime}_{\max}$) values of the projected gradient from the previous training step. Therefore, we do not need to process all the new incoming data points. Considering a set of new data points as $X_\text{new}$, and the subsets ${X'}_\text{new+}$ and ${X'}_\text{new-}$ denoting the positive and negative labeled data respectively. Not necessarily both subsets may exist, \textcolor{black}{and} here we consider that $X_\text{new} ={X'}_\text{new+}$ to maintain the notation. We evaluate the projected gradient of the new set of data points as:
\begin{equation}\label{eq:evaluation}
	\nabla_if(\bm{\alpha_\text{new}})={X^\prime}_\text{new+}\pm\bm{u_+}-1
\end{equation}
This operation \textcolor{black}{keeps} $u_+$ in the coordinate descent procedure of Equation~\ref{eq:u_plus}. We set a new heuristic \textcolor{black}{rule} based on Equation~\ref{eq:DCDshrink} to select only new data points that are more likely to become SVs. We consider the new data points as SVs if the projected gradient values are bounded by ${m^\prime}_{\min}$$<$$\nabla_if(\bm{\alpha_\text{new}})$$<$${m^\prime}_{\max}$. As our method adheres structural and risk minimization principle, all Lagrangian multipliers can be interpreted as SVs, and to let the evaluation of Equation~\ref{eq:evaluation} be more permissive, we can replace the $\max$ and $\min$ operators by the median, mean, or superior and inferior quartiles. Figure~\ref{fig:Inc_FBTSVM} depicts four new data points (in green and numbered), two of each class. The new data points must have projected gradient out of bounds from the respective model to be considered in the incremental procedure. For instance, the circle 1 has a projected gradient lower than ${m^\prime}_{\min}$ of class +1 model, and the circle 2 has a projected gradient greater than ${m^\prime}_{\max}$ of class +1 model. The cross 3 has a projected gradient greater than ${m^\prime}_{\max}$ of class -1, so it is not discarded, but the cross 4 is bounded by the ${m^\prime}_{\min}$ and ${m^\prime}_{\min}$ of class -1 model, so it is discarded. New data points that have projected gradient lower than ${m^\prime}_{\min}$ should interfere in the model shape and placement regarding only its own class, while new data points that have projected gradient greater than ${m^\prime}_{\max}$ interfere in the hyperplane placement regarding the opposite class.
We calculate the membership (as presented in Section~\ref{se:Fuzzy}) to each new data point from $X_\text{new}$ that extrapolates the projected gradient bounds, where $X_\text{over}$ is the data matrix that extrapolates the bounds. Then, we start a new training iteration $k$$\rightarrow$$k+$1 to update the model by enlarging the active set with $X_\text{over}$. Algorithm~\ref{alg:Incremental} presents the pseudo-code for the incremental procedure.     

\begin{algorithm}[htpb!]\footnotesize
	\caption{Incremental procedure}\label{alg:Incremental}\footnotesize
	\textbf{Input:} $X_\text{new}$, $\bm{y}_\text{new}$, $C_1$,$C_2$,$C_3$,$C_4$, and the previous model\\
	\textbf{Output:} updated model 
	\begin{algorithmic}[1]
		\State Let $Z_\text{new-}=\{1,\dots,X_- \}$
		\State Given $\epsilon$, $\bm{\alpha_\text{new}}=0$ and $\bm{u_{+new}} =0$
		\State $\nabla_if(\bm{\alpha}_\text{new})=-H_{-i}\bm{u_+}-1$
		\If{$\nabla_if(\bm{\alpha_\text{new}})>\max m_{\max}$ \text{or} $\nabla_if(\bm{\alpha}_\text{new})<\min m_{\min}$ }\Comment{We check previous $m_{\max}$ and $m_{\min}$'s}
		\State Compute the fuzzy membership $\bm{s}$
		\State Compute $Q=(H^\top_{+} H_++C_1 I)^{-1}H^\top_-$ and ${Q^\prime}_{ii}=H_{-i} Q_i$
		\EndIf
		\While{$ $}
		\For {$\forall i\in Z_\text{new}, $ (a randomly and exclusively selected in the case of batch) } \Comment{this thread runs in parallel}
		\State Run algorithm \ref{alg:Training} from line 8 to 31 
		\EndFor
		\EndWhile
		\State\Return updated model $\leftarrow$ $\bm{u_+}$, $\bm{\alpha}$, $m_{\max}$, $m_{\min}$
	\end{algorithmic}
\end{algorithm}

The incremental procedure adds $X_\text{over}$ data to the model at each iteration, remembering that we need to calculate beforehand the membership value to each data point in $X_\text{over}$, which increases the processing time. In the worst case, we have $X_\text{over}$=$X_\text{new}$, so the model dimension grows linearly with the number of new data points, as well as the processing time increases at each new training iteration. To avoid the continuous growth of the model dimension caused by the incremental procedure, we introduce a decremental procedure to control the model dimension by removing data that has low or no interference in the model accuracy. The proposed decremental procedure is also based on the shrinking technique, where the SVs that have both Lagrangian multipliers smaller than a threshold $(\phi)$ after $(d)$ occurrences are removed. The decremental procedure is executed before each incremental training (except for the first training). We use the vector $Z_{re}$=$[0_1,\dots,0_q]$ (initially all points are assigned to zero) to keep track of the number of occurrences per input data point.  

Considering the current active set $Z$ (without non-bounded Lagrangian multipliers), for each training data point there are two sets of Lagrangian $Z_\alpha$=$\{\alpha_1,\dots,\alpha_{q}\}$ and $Z_\nu$=$\{\nu_1,\dots,\nu_{q}\}$, where $q$ is the number of Lagrangian multipliers. After each training iteration $k\rightarrow k+1$, \textcolor{black}{we update the inputs (in its corresponding position) that results in $(\alpha_m\wedge\beta_m)$$<$$\phi$ in vector $Z_{re}$.} When the number of occurrences reaches $d$, \textcolor{black}{we remove all inputs and related data}, so they will not be used in the next incremental training. Algorithm~\ref{alg:Decremental} presents the pseudo-code for the decremental procedure.    

\begin{algorithm}
	\caption{Decremental procedure}\label{alg:Decremental}\footnotesize
	\textbf{Input:} current model $\rightarrow$ $X$, $\bm{y}$, $\bm{\alpha}$\\
	\textbf{Output:} updated model $\leftarrow$ $X$, $\bm{y}$, $\bm{\alpha}$
	\begin{algorithmic}[1]
		\State Let $Z=\{1,\dots,X \}$ \Comment{We assume that there is an existing classifier}
		\State Given $d$, $\phi$ , $Z_{re}$, $Z_\alpha$, and $Z_\nu$ \Comment{$Z_{re}$ is initially zero}
		\For $\forall i\in Z$
		\If {$Z_{\alpha i}<\phi \wedge Z_{\nu i}<\phi$}
		Increase $Z_{re \; i}$ \EndIf
		\If {$Z_{re \; i}=d$}  
		Remove $Z_{re \; i}, \; Z_{\alpha \; i}, Z_{\nu \; i}$ \EndIf
		\EndFor
		\State\Return updated model $\leftarrow$ $X$, $\bm{y}$, $\bm{\alpha}$
	\end{algorithmic}
\end{algorithm}	
\section{Experimental Results}
\label{se:Experimental}
In this section, we present the experimental protocol\footnote{All tests were performed in a machine running Ubuntu 16.04 LTS with an Intel Core i7-7700HQ CPU @ 2.80GHz and 16,144MB of RAM memory.} used to evaluate the FBTWSVM on benchmarking datasets. For comparison purposes, we have used an experimental protocol similar to Losing et al.~\cite{Losing2018}, \textcolor{black}{which} compares a broad range of state-of-the-art on-line classification algorithms, namely: ISVM with RBF kernel, LASVM with RBF kernel, On-line Random Forest (ORF) \cite{Saffari2009}, Incremental Learning Vector Quantization (ILVQ) \cite{Sato1996}, Learn++ \cite{Polikar2001}, Incremental Extreme Learning Machine (IELM) \cite{Nan-YingLiang2006}, Naive Bayes \cite{Zhang2004}, and Stochastic Gradient Descent (SGD). However, we have restricted to the evaluation of the methods that led to the best accuracy in the on-line learning experiments for at least one of the datasets, which are the ISVM, LASVM, ORF, and ILVQ \cite{Losing2018}. The ORF \cite{Saffari2009} is an incremental Random Forest algorithm that grows continuously from a pre-defined number of trees by adding splits whenever enough data points are gathered within one leaf. It uses Extreme Random Trees \cite{Geurts2006} to optimize the split, using a pre-defined number of random values. The ILVQ is a dynamic growth model derived from the static Generalized Learning Vector Quantization \cite{Sato1996}, where the insertion rate is guided by the number of misclassified data points. The ISVM and the LASVM were already described in Section~\ref{se:FBTSVM}. 

The implementation used for comparison is from \cite{Losing2015a}, \textcolor{black}{which} introduces a prototype placement strategy to minimize the loss of a sliding window of recent data points. The experimental procedure of Losing et al.~\cite{Losing2018} for on-line methods uses a window/chunk size from 500 to 2,000 and set all hyper-parameters using the Hyperopt library \cite{Bergstra2015} with the Tree-of-Parzen-Estimators \cite{Bergstra2011} search algorithm, in which each parameter is individually adjusted within 250 iterations of a 3-fold CV using only the training data. We have carried out all our experiments with FBTWSVM using the approximated RBF kernel described in Section~\ref{sec:kernel}, which enables the use of our linear formulation (Equations~\ref{eq:FRBTSVM1} and \ref{eq:FRBTSVM2}). We optimized the model hyper-parameters using grid-search with a 3-fold cross validation on the training set, and we set the Kernel approximation size following the strategy proposed by Rahimi and Recht \cite{Rahimi2008}.

Considering that we do not need to process all data points to obtain a model, the use of batches accelerates the training phase. In this way, we use different batch sizes but with the constraint that it must encompass at least 5\% of the data points of the fold, and the batch must contain at least one element from each class in the first training. We evaluated the FBTWSVM with six different forgetting window sizes empirically defined as $\varphi$$=$$\{1, 2, 4, 10\}$, and without the decremental procedure. We used publicly available datasets without any preprocessing, although all attributes are numerical, either integer or real values. 
The pre-defined train-tests-splits were used when available. Otherwise, we adopted a stratified train-test-split of 70-30\%.Besides that, we have also created 15 synthetic datasets \cite{Hall2009,Witten2016} to evaluate the scalability of the proposed method as well as a very large dataset of 23M samples \cite{Schmidt2018}. However, for such datasets, we have compared the FBTSVM just with other SVM-based methods. We have used the following streaming generators
with 10\% of noise added: (i) the LED generator \cite{Breiman} yields instances with 24 Boolean features that correspond to the segments of a seven-segment LED display and another 17 irrelevant features; (ii) the SEA generator \cite{Street2001} generates streams from two relevant continuous attributes $f_1, f_2$ and an irrelevant $f_3$, with a range of values within 0 and 10; (iii) the Random Tree Generator (RTG) \cite{Domingos2000} builds a decision tree by randomly selecting attributes as split nodes and assigning random classes to each leaf. 
The number of values per nominal is set to 5, the max tree depth is 3, the first leaf value is 3, and the leaf fraction is 0.15; (iv) the Radial Basis Function (RBF) generator creates 50 centroids at random positions and associates them with a standard deviation value, a weight, and a class label. In this way, new instances are set according to the random direction chosen to offset the centroid, which forms a Gaussian distribution according to the standard deviation associated with the given centroid; (v) HYPER \cite{Domingos2000} generates instances that are separable by a hyperplane. We consider 10\% sigma percentage, and there is no magnitude change or drift attributes. We have created three datasets from each streaming generator with 10,000, 100,000, and 1,000,000 training instances and 3,000, 30,000, and 300,000 testing instances respectively. The focus of our evaluation is in incremental learning considering different key properties (as the number of classes, instances and dimensions), even though we can use the FBTWSVM in offline mode. The datasets\footnote{All datasets and algorithms are available at \url{https://github.com/areeberg/FBTSVM}} encompass generated, artificial and real-world problems with different numbers of classes (from 2 to 100), data points (from 2,586 to 23M) and attributes (from 2 to 5,000), as shown in Table~\ref{table:datasets}, and although the largest dataset has roughly 21 million instances, the proposed system does not specifically target learning from big data.

\addtolength{\tabcolsep}{-2pt}
\begin{table}[t!]
	\centering
	\resizebox{1\hsize}{!}{
		\begin{tabular}{l r r c c c c c c r}
			\hline
			&\multicolumn{2}{c}{\textbf{Number of Examples}}  & \multicolumn{2}{c}{\textbf{Number of}} &  \textbf{Kernel} &  &  &  & \textbf{Number of} \\
			
			\textbf{Dataset}&\multicolumn{1}{c}{\textbf{Train}}  &\multicolumn{1}{c}{\textbf{Test}}  & \multicolumn{1}{c}{\textbf{Attr}} & \multicolumn{1}{c}{\textbf{Classes}} & \multicolumn{1}{c}{\textbf{Size}} & \multicolumn{1}{c}{\textbf{$\boldsymbol{\gamma}$}} & \multicolumn{1}{c}{$\boldsymbol{C_1,C_3}$} & \multicolumn{1}{c}{$\boldsymbol{C_2,C_4}$} & \textbf{Points} \\			\hline
			Border& 4,000 & 1,000 & 2 & 3 & 150                          & 0.4               & 8     & 2     & 100\\ 
			Overlap& 3,960 & 990 & 2 & 4 & 150                          & 0.4               & 8     & 2     & 100  \\ 
			Letter& 16,000 & 4,000 & 16 & 26 & 350                          & 0.01              & 8     & 2     & 1,000 \\ 
			SUSY& 4,500,000 & 500,000 & 18 & 2 & 300                          & 0.2               & 10    & 2     & 100,000  \\ 
			Outdoor& 2,600 & 1,400 & 21 & 40 & 500                          & 0.001             & 10    & 1     & 300 \\ 
			COIL& 1,800 & 5,400 & 21 & 100 & 400                          & 20                & 4     & 4     & 500\\ 
			DNA& 1,400 & 1,186 & 180 & 3 & 500                          & 0.003             & 4     & 4     & 50\\ 
			USPS& 7,291 & 2,007 & 256 & 10 & 1,000                         & 0.007             & 8     & 2     & 1,000 \\ 
			Isolet& 6,238 & 1,559 & 617 & 26 & 1,000                         & 0.002             & 10    & 10    & 500 \\ 
			MNIST & 60,000 & 10,000 & 784 & 10& 2,400                         & 0.0002          & 10    & 10    & 10,000 \\ 
			Gisette & 6,000 & 1,000 & 5,000 & 2 & linear                       & linear            & 8     & 2     & 500 \\
			WESAD & 21,668,504 & 1,537,900 & 8 & 3  & linear                       & linear            & 8     & 2     & 1,537,900 \\
			LED  & 10k\textbar100k\textbar1M & 3k\textbar30k\textbar300k & 24 & 10 & linear                       & linear            & 8     & 2     & 5,000 \\
			SEA & 10k\textbar100k\textbar1M & 3k\textbar30k\textbar300k & 3 & 2 & linear                       & linear            & 10    & 1     & 5,000 \\
			RTG & 10k\textbar100k\textbar1M & 3k\textbar30k\textbar300k & 10 & 2 & 1,400                         & 0.6               & 2.5   & 2     & 5,000\\
			RBF & 10k\textbar100k\textbar1M & 3k\textbar30k\textbar300k & 10 & 5  & 300                          & 0.45              & 8     & 2     & 5,000 \\
			HYPER & 10k\textbar100k\textbar1M & 3k\textbar30k\textbar300k & 10 & 2  & linear                       & linear            & 5     & 4     & 5,000 \\
			\hline
	\end{tabular} }
	\caption{The datasets, their characteristics, and the experimental settings}
	\label{table:datasets}
\end{table}
\addtolength{\tabcolsep}{2pt}

\begin{figure}
	\begin{subfigure}{.5\textwidth}
		\centering
		\includegraphics[width=.99\linewidth]{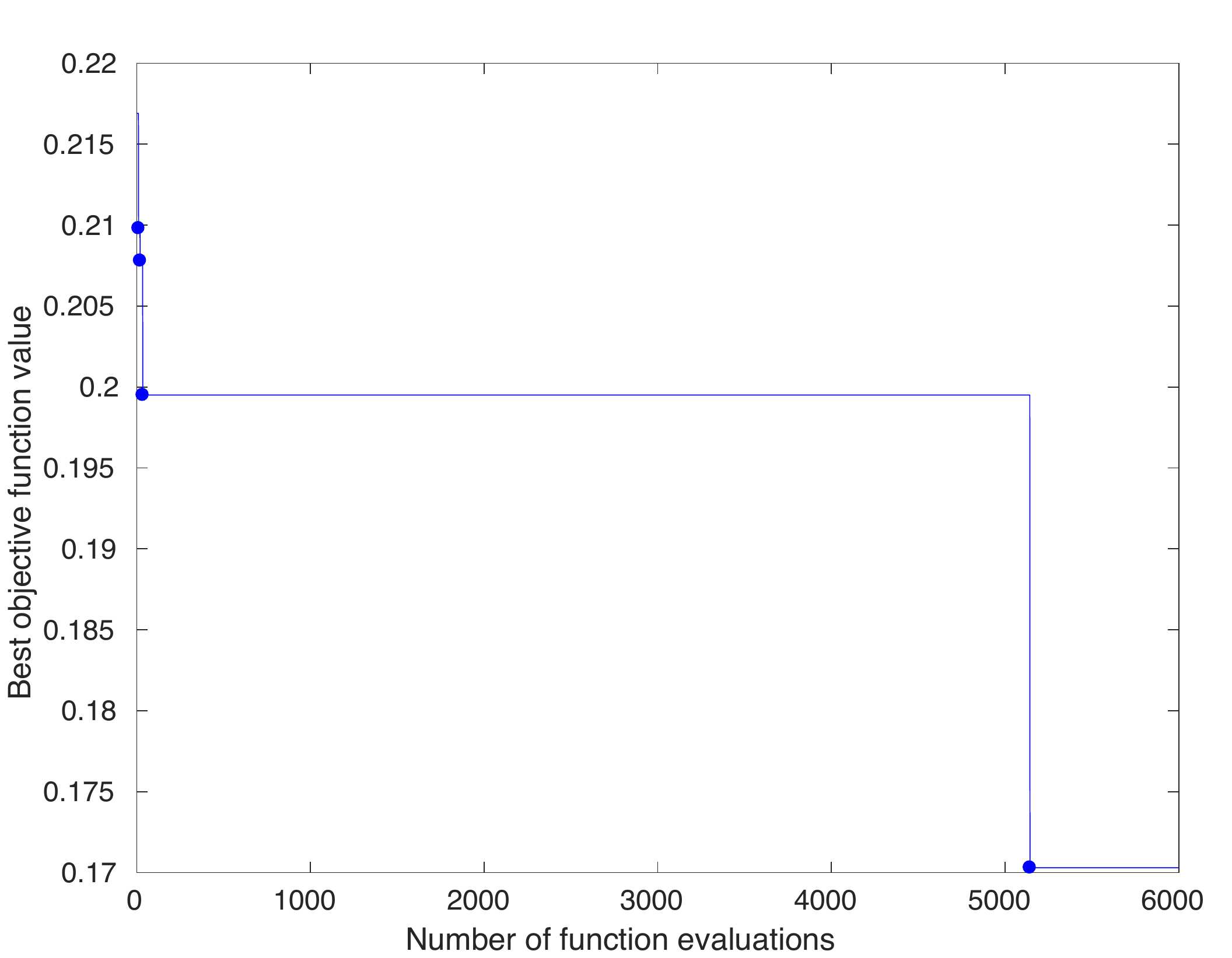}
		\caption{Convergence plot for the 4-D case.}
		\label{fig:sfig1}
	\end{subfigure}%
	\begin{subfigure}{.5\textwidth}
		\centering
		\includegraphics[width=.99\linewidth]{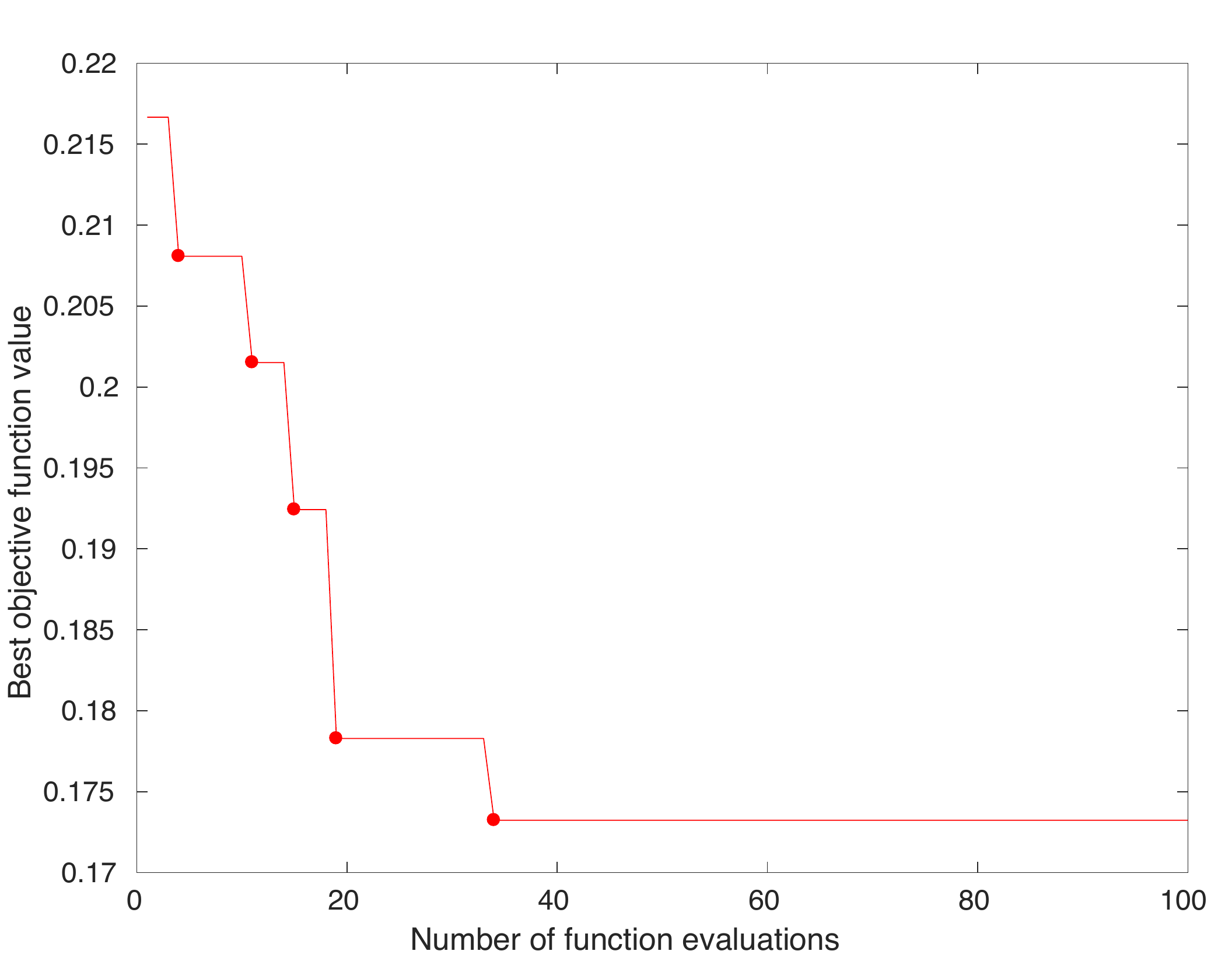}
		\caption{Convergence plot for the 2-D case.}
		\label{fig:sfig2}
	\end{subfigure}
	\caption{Overlap dataset convergence plot considering the 4-D and 2-D cases.}
	\label{fig:Convergence}
\end{figure}

Table~\ref{table:datasets} also shows the parameter setting used for each dataset which was defined in a 3-fold CV, and "Number of Points" stands for the initial training set size. For all datasets we used a fixed fuzzy parameter $ \mu =0.1$. Using the 4-D case ($C_1, C_2, C_3, C_4$ are independent variables) for the hyperparameter tuning may result in a model with a better generalization performance, i.e., the loss function may achieve a lower value during the model selection compared to the 2-D case (we assume $C_1$=$C_3$ and $C_2$=$C_4$), however, performing the hyperparameter tuning in a 2-D space may decrease substantially the number of function evaluations needed, especially given that the grid search is essentially a brute force search strategy that takes long. Other model selection strategies are able to speed-up the hyperparameter tuning, however, this is out of the scope of this paper \cite{Mello2019}. In many cases, using the 2-D space instead of the 4-D is a valid heuristic estimation to decrease the number of function evaluations needed, and using the Overlap dataset as an example, Figure~\ref{fig:Convergence} depicts that using the 2-D space it requires 34 function evaluations to achieve the accuracy loss value of 0.1732, while the 4-D space requires 5,143 function evaluations to achieve 0.1703.

Table~\ref{table:accuracy} \textcolor{black}{presents} the accuracy of the incremental and decremental FBTWSVM against the best on-line algorithms reported in \cite{Losing2018}. The FBTWSVM achieved equal or better results in 8 out of 11 datasets (from Border to Gisette, excluding the generated datasets) relative to the best on-line algorithms. The SUSY dataset contains a significant amount of data and to train the FBTWSVM we had limited the size of the kernel approximation based on the memory available, and this also reduces the accuracy. For instance, both the ISVM and the LASVM with an RBF kernel could not be trained with this dataset due to the uncontrolled growth of the kernel matrix. We run an experiment using the SUSY full training set considering a kernel approximation size of 600, resulting in 77.67\%. Outdoor is a visual dataset that consists of objects recorded outdoors under lighting conditions \cite{Losing2015a}. The dataset creation method has caused a difference between training and test data \cite{Losing2018}, which reflects on the performance of the learning algorithms. On-line algorithms with an adaptive learning mechanism presented the accuracy of about 20\% better than off-line methods (the best result found was the off-line ISVM with 71.9\% \cite{Losing2018}). 

\begin{table}
	\centering
	\caption{On-line accuracy of the incremental and decremental FBTWSVM compared to other incremental algorithms on several benchmark datasets.  Statistically significant differences are marked with $^\star$.}
	\label{table:accuracy}
	\resizebox{0.75\hsize}{!}{
		\begin{tabular}{lllllrl}
			\hline
			\multicolumn{1}{l}{} & \multicolumn{6}{c}{\textbf{Accuracy (\%)} }  \\ 
			\cline{2-7}
			\textbf{Dataset}    & \multicolumn{2}{c}{\textbf{FBTWSVM} }          & \multirow{2}{*}{\begin{tabular}[c]{@{}l@{}}\textbf{ISVM} \\ \end{tabular}} & \multirow{2}{*}{\textbf{LASVM} } & \multicolumn{1}{l}{\multirow{2}{*}{\textbf{ORF} }} & \multirow{2}{*}{\textbf{ILVQ} }  \\ 
			\cline{2-3}
			\textbf{}           & \textbf{Best}           & \textbf{Mean$\pm$SD}  &                                                                            &                                  &                            &                                  \\ 
			\hline
			Border               & \textbf{98.70}          & 97.60$\pm$1.10            & 98.50                                                                      & 97.6                             & 94.0                                                & 94.7                             \\ 
			Overlap              & \textbf{84.14$^\star$}  & 82.58$\pm$1.49            & 81.7                                                                       & 78.8                             & 78.2                                                & 81.1                             \\ 	
			Letter               & \textbf{96.75$^\star$}  & 96.68$\pm$0.07            & 91.3                                                                       & 92.7                             & 75.4                                                & 88.4                             \\ 
			SUSY                 & 77.67                   & 76.00$\pm$1.20            & -                                                                          & -                                & \textbf{79.3}                                       & 78.5                             \\ 
			Outdoor              & 74.44                   & 73.72$\pm$0.42            & \textbf{86.4}                                                              & 82.3                             & \begin{tabular}[c]{@{}r@{}}34.2\\ \end{tabular}     & 82.6                             \\ 
			COIL                 & \textbf{95.11$^\star$}  & 94.99$\pm$0.14            & 75.4                                                                       & 66.3                             & 66.6                                                & 79.1                             \\ 
			DNA                  & \textbf{93.59$^\star$}  & 92.90$\pm$0.30            & 89.5                                                                       & 89.5                             & 73.1                                                & 84.6                             \\ 
			USPS                 & 95.47                   & 94.91$\pm$0.30            & \textbf{96.7}                                                              & 96.6                             & \begin{tabular}[c]{@{}r@{}}84.5\\ \end{tabular}     & 92.7                             \\ 
			Isolet               & \textbf{96.28$^\star$}  & 95.88$\pm$0.37            & 93.6                                                                       & 92.9                             & 69.2                                                & 84.7                             \\ 
			MNIST                & \textbf{97.80}          & 97.00$\pm$0.12            & -                                                                          & 97.5                             & 87.1                                                & 90.8                             \\ 
			Gisette              & \textbf{96.50}          & 96.40$\pm$0.01            & 96.3                                                                       & 96.4                             & 90.3                                                & 91.1\\
			\hline
			\multicolumn{7}{l}{\color{black}" - " denotes the non-available results due to limitations in memory size.}
	\end{tabular}}
\end{table}

Table~\ref{table:accuracy_SV} shows the relation between accuracy and the number of SVs resulting from different forgetting scores ($d$). The decremental procedure discards points that are less likely to be SVs. Smaller $d$ leads to classifiers with lower generalization performance, and for most of the datasets, the best performance was achieved without forgetting or with large forgetting scores. On the other hand, the number of SVs using the decremental procedure is considerably smaller, so the forgetting score must be chosen according to the application. Table~\ref{table:accuracy_SV} also presents the comparison between the online and offline approaches, in which the online had better accuracy to all datasets with a smaller number of SVs compared to offline. A smaller $d$ also implies in a faster training and classification time, and Table~\ref{table:time_IFBTSVM} shows that the difference in training time can be substantial (check the SUSY dataset values for example). The accuracy of the COIL dataset with a forgetting score $d$=10 have similar accuracy (95.11\%) when compared to the offline implementation of the FBTWSVM (95.00\%), the ISVM (96.50\%), the  LASVM (93.20\%) \cite{Losing2018}, and the multiclass SVM implemented with the Error-Correcting Output Codes (ECOC) from MATLAB (96.52\%). In this manner, the forgetting strategy does not discard crucial support vectors, keeping the accuracy score near the offline approach. Table~\ref{table:time_IFBTSVM} presents the accuracy performance evolution when increasing the forgetting score, which corroborates with the forgetting strategy, i.e., lower forgetting scores tend to have smaller accuracy, however, by keeping the important SV the accuracy does not fall substantially (the accuracy difference between $d$=1 and $d$=10 is 1.9\%). 

\addtolength{\tabcolsep}{-2pt}    
\begin{table*}[htpb!]
	\scriptsize
	\centering
	\caption{On-line accuracy with different forgetting scores ($d$) and the corresponding number of support vectors (nSV).}
	\label{table:accuracy_SV}
	\resizebox{1\hsize}{!}{
		\begin{tabular}{l l l l l l l}
			\hline
			 & \multicolumn{6}{c}{\textbf{Accuracy(\%)}}\\ 
			\textbf{Dataset}& \multicolumn{1}{l}{\textbf{$d$=1\hspace{2pt}$|$nSV}} & \multicolumn{1}{l}{\textbf{$d$=2 \hspace{2pt}$|$nSV}}  & \multicolumn{1}{l}{\textbf{$d$=4\hspace{4.7pt}$|$nSV}}  & \multicolumn{1}{l}{\textbf{$d$=10\hspace{0.7pt}$|$nSV}} & \multicolumn{1}{l}{\textbf{$d$=$\infty$\hspace{0.9pt}$|$nSV}} & \multicolumn{1}{l}{\textbf{OffL$|$nSV}} \\ 
			\hline
			Border & 91.90$|$538 & 93.30\hspace{0.2pt} $|$784 & 98.20\hspace{0.2pt} $|$1.3k & \hspace{0.2pt}97.20$|$2.8k & \textbf{98.70$|$7k} & 98.50$|$8k \\ 
			Overlap & 76.97$|$1.9k & 79.70\hspace{0.2pt}  $|$2.3k  & 82.22\hspace{0.2pt} $|$3.5k & \hspace{0.2pt}82.83$|$7k & \textbf{84.14$|$11.8k} & 83.30$|$11k\\ 
			Letter & 93.38$|$83k & 94.83\hspace{0.2pt} $|$122k  & 96.10\hspace{0.2pt} $|$204k & \hspace{0.2pt}96.10$|$338k & \textbf{96.63$|$361k} & 96.90$|$384k\\ 
			SUSY & 45.84$|$754k & 45.85\hspace{0.2pt} $|$1.5M  & \textbf{77.67$|$2.5M} & \hspace{0.2pt}73.78$|$3.4M & \hspace{3.4pt}76.45$|$3.8M & - \\ 
			Outdoor & 72.00$|$24k & 72.25\hspace{0.2pt} $|$39k & 73.69\hspace{0.2pt} $|$68k &  \textbf{74.44$|$83k} & \hspace{3.4pt}73.88$|$92k & 74.00$|$93k \\ 
			COIL & 93.21$|$98k & 94.01\hspace{0.2pt} $|$100k & 94.57\hspace{0.2pt} $|$153k & \textbf{95.11$|$156k} &\hspace{3.4pt}94.93$|$166k &95.00$|$178k \\ 
			DNA & 91.82$|$770 & 92.16\hspace{0.2pt} $|$848 & 92.50\hspace{0.2pt} $|$1.2k & \textbf{93.78$|$1.9k} & \hspace{3.4pt}93.59$|$2.6k & 93.50$|$2.7k\\ 
			USPS & 93.92$|$9k & 94.32\hspace{0.2pt} $|$21k & 94.87\hspace{0.2pt} $|$42k  & \hspace{0.2pt}95.36$|$60k & \textbf{95.47$|$60k} & 95.30$|$65.5k\\ 
			Isolet & 95.19$|$61k & 95.51\hspace{0.2pt} $|$85k & 95.32\hspace{0.2pt} $|$128k &  \hspace{0.2pt}95.89$|$143k & \textbf{96.28$|$143k} & 95.60$|$155k\\ 
			MNIST & 97.17$|$59k & 97.48\hspace{2.5pt}$|$169k & 97.66\hspace{0.2pt} $|$342k & \textbf{97.91$|$455k} & \hspace{3.4pt}97.80$|$455k & 97.80$|$540k\\ 
			Gisette & 96.50$|$1.3k & \textbf{97.00$|$1.9k}  & 96.90\hspace{3.5pt}$|$2.9k &  \hspace{0.2pt}96.20$|$5.3k & \hspace{3.4pt}96.50$|$6k & 96.50$|$6k\\ 
			\hline
		\multicolumn{7}{l}{\color{black}" - " denotes the non-available results due to limitations in memory size.}
	\end{tabular}}
\end{table*}
\addtolength{\tabcolsep}{9pt}
\begin{table}[htpb!]
	\scriptsize
	\centering
	\caption{Training and testing processing time with different forgetting scores ($d$) in seconds.}
	\label{table:time_IFBTSVM}
	\resizebox{0.85\hsize}{!}{
		\begin{tabular}{l r r r}
			\hline
		&\multicolumn{3}{c}{\textbf{Training  $|$ Testing Time (sec)}}  \\ \cline{2-4}
			\textbf{Dataset} & \multicolumn{1}{c}{\textbf{$d$=1}}  & \multicolumn{1}{c}{\textbf{$d$=10}} & \multicolumn{1}{c}{\textbf{$d$=$\infty$}} \\ \hline
			Border  & \ \ 12.39  $|$ 0.01  & 34.12 $|$ 0.02   & \ \ 43.22 $|$ 0.02   \\ 
			Overlap & \ \ 22.96  $|$ 0.02  & 82.58 $|$ 0.02   & \ \ 94.37 $|$ 0.02   \\ 
			Letter  & \ \ 88.14  $|$ 0.68  & 179.18 $|$ 0.72  & 192.36 $|$ 0.70  \\ 
			SUSY    & 940.60 $|$ 1.88 & 5264.41 $|$ 2.62 & {-}\hspace{6pt} $|$\hspace{6pt}{-}\hspace{8pt}           \\ 
			Outdoor & 111.00 $|$ 0.73 &  \ \ 153.09 $|$ 0.80  & 153.55 $|$ 0.81  \\ 
			COIL    & 154.63 $|$ 5.00 &  \ \ 169.10 $|$ 5.29  & 179.12 $|$ 4.51  \\ 
			DNA     & \ \ \ \ 6.89  $|$ 0.03   & 8.63 $|$ 0.03    & \ \ \ \ 8.53 $|$ 0.03    \\ 
			USPS    & \ \ 21.59 $|$ 0.25  &  41.59 $|$ 0.24   & \ \ 41.67 $|$ 0.23   \\ 
			Isolet  & 117.35 $|$ 0.52 &  \ \ 181.14 $|$ 0.49  & 160.10 $|$ 0.51  \\ 
			MNIST   & 349.47 $|$ 1.21 &  \ \ 419.49 $|$ 1.19  & 435.48 $|$ 1.26  \\ 
			Gisette & \ \ 25.26 $|$ 0.01  &   50.52 $|$ 0.01   & \ \ 49.41 $|$ 0.01  \\
			\hline
		\multicolumn{4}{l}{\color{black}" - " denotes the non-available results due to limitations in memory size.}
	\end{tabular}}
\end{table}

Table~\ref{table:synthetic} compares the training time (in seconds), the real RAM consumed of the current process and its children (in Gigabytes), and the accuracy of the FBTSVM with the other SVM based methods (ISVM and LASVM). For these experiments we split the dataset into the largest batches that we can (that fits on the available memory, initially 15.4 Gb), to reduce the reloading procedure of the dataset during the execution (more loading implies in a larger training time). Both the FBTWSVM and the ISVM (the ISVM multiclass adopt one-versus-one strategy) are implemented in MATLAB, thus it requires more real RAM than the LASVM, that is for binary cases only and it is a C++ implementation. We do not consider the LASVM in the multiclass cases (LED and RBF), and we discard the situations that the training time took over 12 hours. All methods present competitive accuracy, however, the FBTSVM is the only method (compared to ISVM and LASVM) able to train all dataset sizes in an acceptable time, having the smallest training time for almost all situations (the only exception is the LASVM for the RTG10K). The FBTSVM forgetting strategy is one of the factors (the kernel approximation also plays an important role) that makes the training into large datasets possible, as Table~\ref{table:synthetic} shows that the real RAM consumed difference between the 100K and 1M datasets is not very expressive. The LED dataset has a bigger memory difference between the datasets for the FBTWSVM, and this is caused by the use of the multi-thread instead of the single processor version. 
In this way, the scalability of the FBTWSVM is superior to  other online SVM-based methods, as it requires a smaller training time to process large datasets and can handle the memory consumption in an efficient manner. To further explore the FBTWSVM potential for large datasets, we have also evaluated the accuracy, training time, and memory consumption on the WESAD dataset \cite{Schmidt2018} considering three classes (baseline, stress, and amusement), eight attributes acquired from a sensor attached to the chest, and using the leave-one-subject-out cross-validation (in total we have 17 subjects). The best result reported by Schmidt et al.~\cite{Schmidt2018} is 76.50\% using a Linear Discriminant Analysis, however, this is an offline approach and the authors do not present the training time or memory consumption. Our method achieved the accuracy of 75.50\% (Table~\ref{table:synthetic}), with a training time of 6,789 seconds and peak memory consumption of 9.8 GB.

\addtolength{\tabcolsep}{-10pt}
\begin{table*}[htpb!]
{\color{black}
	\scriptsize
	\caption{Comparison of the training time (TT) in seconds (s), real memory usage (RMU) in Gigabytes (GB) and percentual accuracy (Acc) (\%) for five synthetic datasets and three different amount of data, and WESAD dataset.}
	\label{table:synthetic}
	\renewcommand{\arraystretch}{1.2}
	\resizebox{1\hsize}{!}{
		\begin{tabular}{l l c c c c c c }
			\hline
			& & \multicolumn{6}{c}{Dataset} \\
			& & \multicolumn{6}{c}{10k$|$100k$|$1M} \\
			\cline{3-7}
			\multicolumn{1}{c}{Method} & & LED & SEA & RTG & RBF & HYPER & WESAD \\
			\hline
			\multirow{3}{*}{FBTWSVM} & TT & 5.81$|$19.1$|$380\hspace{3pt} & 0.69$|$2.15$|$63\hspace{6pt} & 9.60$|$\hspace{1pt}119\hspace{1pt}$|$3.4k & 11.4$|$\hspace{1pt}164\hspace{1pt}$|$5.3k & 0.73$|$2.61$|$28.0 & \multicolumn{1}{c}{6.8k}\\
			 & RMU & 3.26$|$3.81$|$7.34 & 0.84$|$0.91$|$1.31 & 1.84$|$10.1$|$10.6 & 1.04$|$2.06$|$7.03 & 0.84$|$0.96$|$1.47 & \multicolumn{1}{c}{9.80}\\
			& Acc & 74.1$|$74.1$|$74.2 & 89.0$|$87.1$|$89.1 & 95.8$|$95.5$|$96.0 & 88.3$|$89.4$|$89.0 & 89.1$|$94.7$|$94.0 & \multicolumn{1}{c}{75.5}\\
			\hline
			\multirow{3}{*}{ISVM} & TT & 40.7$|$\hspace{5pt}-\hspace{5pt}$|$\hspace{5pt}-\hspace{5pt} & 8.80$|$4.5k$|$\hspace{5pt}-\hspace{5pt} & 63.9$|$\hspace{5pt}-\hspace{5pt}$|$\hspace{5pt}-\hspace{5pt} & 31.8$|$\hspace{5pt}-\hspace{5pt}$|$\hspace{5pt}-\hspace{5pt} & 14.4$|$4.9k$|$\hspace{5pt}-\hspace{5pt} &\multicolumn{1}{c}{-}\\
			 & RMU &  0.89$|$\hspace{5pt}-\hspace{5pt}$|$\hspace{5pt}-\hspace{5pt} & 0.88$|$0.98$|$\hspace{5pt}-\hspace{5pt} & 0.94$|$\hspace{5pt}-\hspace{5pt}$|$\hspace{5pt}-\hspace{5pt} & 0.94$|$\hspace{5pt}-\hspace{5pt}$|$\hspace{5pt}-\hspace{5pt} & 0.82$|$0.95$|$\hspace{5pt}-\hspace{5pt} &\multicolumn{1}{c}{-}\\
			& Acc & 74.2$|$\hspace{5pt}-\hspace{5pt}$|$\hspace{5pt}-\hspace{5pt} & 89.9$|$89.3$|$\hspace{5pt}-\hspace{5pt} & 90.0$|$\hspace{5pt}-\hspace{5pt}$|$\hspace{5pt}-\hspace{5pt} & 87.6$|$\hspace{5pt}-\hspace{5pt}$|$\hspace{5pt}-\hspace{5pt} & 94.1$|$93.0$|$\hspace{5pt}-\hspace{5pt} &\multicolumn{1}{c}{-}\\
			\hline
			\multirow{3}{*}{LASVM} & TT & \multicolumn{1}{c}{-} & 2.44$|$\hspace{1pt}328\hspace{1pt}$|$\hspace{5pt}-\hspace{5pt} & 2.39$|$\hspace{1pt}546\hspace{1pt}$|$\hspace{5pt}-\hspace{5pt} & \multicolumn{1}{c}{-} & 2.21$|$\hspace{1pt}519\hspace{1pt}$|$\hspace{5pt}-\hspace{5pt} &\multicolumn{1}{c}{-}  \\
			& RMU & \multicolumn{1}{c}{-} & 0.06$|$0.36$|$\hspace{5pt}-\hspace{5pt} & 0.03$|$0.41$|$\hspace{5pt}-\hspace{5pt} & \multicolumn{1}{c}{-} & 0.05$|$0.37$|$\hspace{5pt}-\hspace{5pt} &\multicolumn{1}{c}{-}\\
			& Acc & \multicolumn{1}{c}{-} & 84.0$|$87.0$|$\hspace{5pt}-\hspace{5pt} & 88.0$|$95.7$|$\hspace{5pt}-\hspace{5pt} & \multicolumn{1}{c}{-} & 91.7$|$93.8$|$\hspace{5pt}-\hspace{5pt} &\multicolumn{1}{c}{-}\\
			\hline
			\multicolumn{8}{l}{" - " denotes the non-available results due to training times greater than 12 hours.} 
	\end{tabular}}
	}
\end{table*}
\addtolength{\tabcolsep}{3pt}

\section{Conclusion}
\label{se:Conclusion}
In this paper, we propose a novel SVM approach suitable for incremental and decremental on-line learning. The incremental and decremental Fuzzy Bounded Twin SVM (FBTWSVM) integrates ideas coming from different SVM approaches such as the Twin SVM \cite{Jayadeva2007}, the Fuzzy SVM \cite{Chun-FuLin2002}, the Bounded TWSVM \cite{Yuan-HaiShao2011}, the Fast and Robust TWSVM \cite{Gao2015,Gao2017}, the Optimal DAG TWSVM \cite{JingChen2010}, and the dual coordinate descent method \cite{Chang2008}. The FBTWSVM calculates a pair of non-parallel hyperplanes using two smaller QPPs, rather than one large QPP as in the original SVM, but with adherence to structural risk minimization principle. The dual form of the FBTWSVM leads to a pair of convex quadratic programming problems with a unique solution and singularity avoidance. The dual coordinate descent method with shrinking requires less memory storage than the TWSVM, as it discards points that are less likely to be SVs. The fuzzy concept enhances noise-resistance and generalization capability, while the use of a kernel approximation shows a good generalization performance with our linear model.

The incremental solution follows the shrinking strategy and can run with different batch sizes, from a single individual to a number of data points that fits the available memory. The decremental procedure is fundamental to control the model complexity, keeping only the most critical SVs in the model. The FBTWSVM is flexible and both incremental and decremental procedures can be configured according to the application, changing the threshold of adding new SVs in the incremental step and the number of occurrences in the decremental step. According to the experimental results, the DAG strategy showed a good generalization capability and a fast training speed, but for further studies the use of training data structural and statistical information in the training process may increase the generalization performance. A practical difficulty in the FBTWSVM is the optimization of the six hyper-parameters $C_1, C_2, C_3, C_4, \mu, \gamma$ and the kernel approximation size, however, this problem will be addressed in the future. 
The FBTWSVM can adapt current models using the window strategy, or even add new models (e.g. in case of new classes) without retraining. Therefore, as a future work, we will evaluate the FBTWSVM use in the context of concept drift, novelty detection, and big data.  

\section*{}
\singlespacing
\bibliographystyle{model1-num-names}


\end{document}